\documentclass{article}

\usepackage{microtype}
\usepackage{graphicx}
\usepackage{booktabs}
\usepackage{hyperref}
\usepackage{subcaption}
\usepackage{amsmath}
\usepackage{amsthm}
\usepackage{amsfonts}
\usepackage{amssymb}
\usepackage{adjustbox}
\usepackage{multicol}
\usepackage{mathtools}
\usepackage[shortlabels]{enumitem}
\usepackage{bm}

\newcommand{\R}{\mathbb{R}}

\newcommand{\ex}{\mathbb{E}}

\newtheorem{proposition}{Proposition}

\setlist[itemize]{align=parleft,left=1em..2em}
\setlist[enumerate]{align=parleft,left=0.5em..2em}

\usepackage[accepted]{mlsys2025}

\mlsystitlerunning{Interference-Aware Edge Runtime Prediction with Conformal Matrix Completion}

\begin{document}

\twocolumn[
\mlsystitle{Interference-Aware Edge Runtime Prediction with Conformal Matrix Completion}




\begin{mlsysauthorlist}
\mlsysauthor{Tianshu Huang}{cmu}
\mlsysauthor{Arjun Ramesh}{cmu}
\mlsysauthor{Emily Ruppel}{bosch} \\
\mlsysauthor{Nuno Pereira}{isep}
\mlsysauthor{Anthony Rowe}{cmu,bosch}
\mlsysauthor{Carlee Joe-Wong}{cmu}
\end{mlsysauthorlist}

\mlsysaffiliation{cmu}{Department of Electrical and Computer Engineering, Carnegie Mellon University}
\mlsysaffiliation{bosch}{Robert Bosch GmbH}
\mlsysaffiliation{isep}{School of Engineering (ISEP/IPP) and INESC TEC, Polytechnic Institute of Porto}

\mlsyscorrespondingauthor{Tianshu Huang}{tianshu2@andrew.cmu.edu}

\mlsyskeywords{runtime prediction, execution time prediction, performance analysis, matrix factorization, matrix completion, webassembly, bytecode analysis, uncertainty quantification, conformal prediction, quantile regression, machine learning}

\vskip 0.3in

\begin{abstract}
Accurately estimating workload runtime is a longstanding goal in computer systems, and plays a key role in efficient resource provisioning, latency minimization, and various other system management tasks. Runtime prediction is particularly important for managing increasingly complex distributed systems in which more sophisticated processing is pushed to the edge in search of better latency. Previous approaches for runtime prediction in edge systems suffer from poor data efficiency or require intensive instrumentation; these challenges are compounded in heterogeneous edge computing environments, where historical runtime data may be sparsely available and instrumentation is often challenging. Moreover, edge computing environments often feature multi-tenancy due to limited resources at the network edge, potentially leading to interference between workloads and further complicating the runtime prediction problem. Drawing from insights across machine learning and computer systems, we design a matrix factorization-inspired method that generates accurate interference-aware predictions with tight provably-guaranteed uncertainty bounds. We validate our method on a novel WebAssembly runtime dataset collected from 24 unique devices, achieving a prediction error of 5.2\% -- 2x better than a naive application of existing methods.
\end{abstract}
]



\printAffiliationsAndNotice{}  

\section{Introduction}

Practitioners have always had to balance hardware capacity with expected software resource demands, for example, when specifying future hardware platforms or optimizing application placement across cloud datacenter servers based on predicted application runtimes \citep{amiri2017survey}. This task has only been made more difficult as improvements in compute and networking capabilities drive a paradigm shift pushing compute to \textit{edge systems} closer to the physical world such as embedded computers and smart phones.

Edge systems present a unique set of challenges. Unlike cloud systems, edge systems are highly heterogeneous, involving compute platforms across a wide range of compute capabilities ranging from microcontrollers to edge servers. Edge systems are also highly resource constrained, and can suffer greatly from interference between workloads \cite{cavicchioli2017memory}, yet benefit greatly from multi-tenancy since the alternative --- cloud offloading --- may result in unnacceptable network latency. In addition to these complexities, edge computing applications also often impose additional quality of service requirements such as deadlines and latency constraints, which must be balanced with the availability of compute, networking, or even power \citep{satyanarayanan1996fundamental, zambonelli2004spatial}.


The latency-sensitive nature of many edge workloads makes it useful, and sometimes crucial, to anticipate workload runtimes in deployment to ensure that they meet their quality-of-service needs. For example, an industrial controller on a manufacturing line may need to complete within a given timeframe with high probability to forestall interruptions to the manufacturing plant, or a smartphone might need to decide which model to load for a particular inference task. Indeed, such runtime performance measures are crucial for edge orchestration frameworks that aim to ensure workload performance by placing them on different available platforms. However, the heterogeneous, resource-limited nature of edge systems also limits the practicality of exhaustively benchmarking all possible deployments: even if each workload can be benchmarked on each platform, all combinations of potentially interfering workloads cannot.

In the absence of benchmark data, orchestrators must instead \textit{predict} workload performance, of which predicting the execution time, or \textit{runtime}, is a critical component, especially with latency-sensitive workloads. As a further complication, accurate runtime estimates may not always be possible: accuracy may be constrained by insufficient data, or, since edge devices are harder to maintain than cloud datacenters, even unpredictable variations in the platforms themselves. Thus, we focus on three key challenges: maximizing data efficiency for a \textit{matrix completion} formulation, \textit{predicting interference} between concurrent workloads, and \textit{quantifying the uncertainty} of our predictions.

\paragraph{Matrix Completion}

Predicting workload runtime across heterogeneous platforms must predict the impact of complex effects such as differences in compiler optimizations \citep{hoste2008cole}, operating system performance \citep{chen1993impact}, and hardware architecture \citep{zheng2015learning}. Generalizing runtime predictions from a limited number of observations to multiple heterogeneous platforms thus requires us to intelligently combine observations of how different workloads perform on various platforms. While matrix completion is not new to performance analysis, we are the first to use it for explicit runtime prediction.

\paragraph{Interference Modeling}

\begin{figure}
    \centering
    \includegraphics[width=\columnwidth]{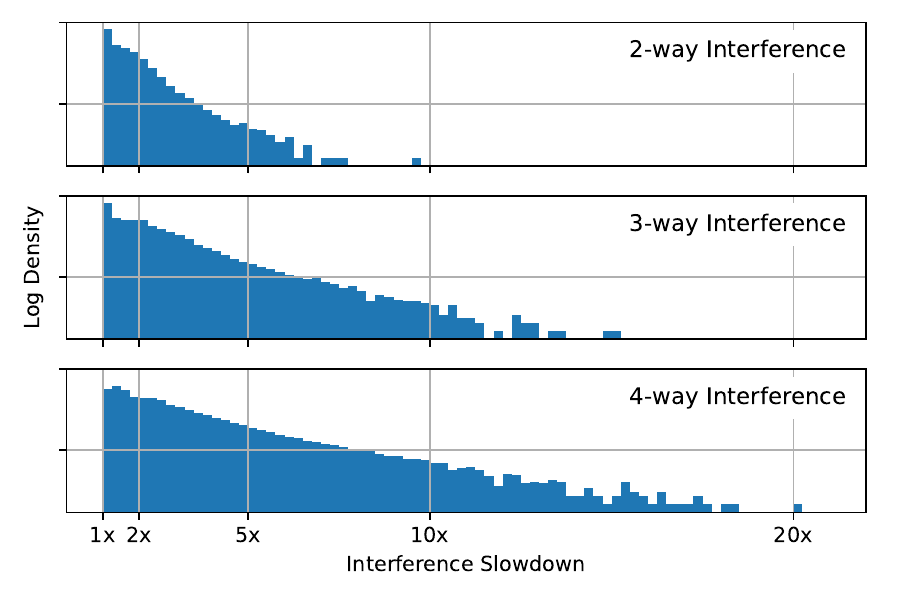}
    \vspace{-2.5em}
    \caption{Log-histogram of interference effects in our dataset, sorted by the number of interfering workloads; we observe up to a 20$\times$ slowdown in randomly sampled benchmark combinations.}
    \label{fig:interference_histogram}
\end{figure}

Workloads running simultaneously (Fig.~\ref{fig:interference_histogram}) can cause a range of interfering effects due to shared resource contention \cite{chai2007understanding, cavicchioli2017memory}. This is of particular concern in edge systems, where co-locating interfering workloads may be preferable to the latency of offloading to a remote server.

Many prior works capture set interference types using micro-benchmarks \cite{delimitrou2013paragon, delimitrou2014quasar}; this strategy is well-suited for cloud systems with well-defined resources such as CPU, memory, and networking. However, edge systems can suffer from difficult-to-quantify types of resource contention such as scheduling and alignment, power and clock speed effects, or even contention for application-specific communication channels such as Bluetooth or CAN. We instead focus on an \textit{observation-based} interference-aware prediction method that learns interference patterns by simply observing workload behavior, using a novel extension of matrix factorization.

\paragraph{Uncertainty Quantification}

Not all workloads and platforms have performance characteristics which are equally predictable: some may be more susceptible to indeterminism, have more complex behavior, or simply have less training data available. Thus, while a ``mean'' runtime prediction can be useful, an estimate which captures the \textit{uncertainty} of a model's prediction can better guide design and deployment decisions --- whether for an engineer or for an orchestration algorithm \cite{alipourfard2017cherrypick}.

We formalize this requirement by predicting \textit{runtime bounds}: for a given tail $\varepsilon$, what runtime budget will be sufficient for a workload to complete with probability at least $1 - \varepsilon$? Applying conformal prediction \cite{shafer2008tutorial}, we develop a quantile regression-based procedure \cite{romano2019conformalized} for predicting tight bounds in probability.

\paragraph{Contributions} Our runtime prediction method, Pitot\footnote{
Our code and dataset are open source, and can be found at \url{https://github.com/wiseLabCMU/pitot}; an archival copy is also available at
\url{https://zenodo.org/records/14977004}.
}, integrates a collection of novel contributions that draw insight across machine learning and systems theory, and provides up to a 2x improvement over a naive combination of existing methods. To summarize our contributions:
\begin{enumerate}[(1), topsep=0pt, itemsep=0pt]
    \item \textbf{Runtime Prediction for the Edge}: We make several novel contributions to the matrix completion method (Sec.~\ref{sec:method}) that address challenges found in edge systems, including a \textit{log-residual objective} to handle high heterogeneity, a \textit{matrix factorization with side information} formulation to exploit workload and platform features, an \textit{interference prediction} term to predict workload interference, and \textit{uncertainty quantification} to probabilistically bound runtime.

    \item \textbf{Runtime Dataset}: In order to develop and evaluate our method, we collect a novel dataset with 410,970 unique data points from 249 benchmarks, 10 different runtime configurations, and 24 devices.
    Our dataset uses WebAssembly \cite{haas2017bringing}, a lightweight virtualization framework which allows us to run common benchmarks on platforms ranging from x86 servers to microcontrollers. WebAssembly has also seen increasing adoption from the cloud \cite{fastly-wasm, cloudflare-wasm, shopify-wasm} to edge applications such as IoT \cite{li2022bringing}, augmented reality \cite{pereira2021arena}, automotive \cite{etas-wasm, scheidl2020webassembly}, and industrial automation \cite{siemens-wasm}.
    
    \item \textbf{Evaluation}: We run extensive experiments demonstrating the impact of each of our methodological contributions and verifying the efficacy of our method. In particular, Pitot achieves as low as 5.2\% error, and can generate tightly-bounded prediction intervals (Sec.~\ref{sec:results}).
\end{enumerate}

\section{Related Work}
\label{sec:related}

\paragraph{Worst-Case Execution Time}

Predicting the worst-case execution time (WCET) --- the longest time a program takes to finish, considering all possible inputs --- has strong roots in real-time and embedded systems \citep{wilhelm2008worst}. WCET algorithms bound runtime execution using pessimistic assumptions on aspects of program execution such as memory access \citep{ferdinand2007static}, loop counts \citep{puschner1997computing}, thread interaction \citep{syswcet}, and execution paths \citep{ermedahl1997deriving}. However, WCET estimates of ordinary programs can be extremely large or even unbounded, leading to the need for more data-driven runtime prediction approaches.

\paragraph{Platform Modeling}

Learning a performance model for each platform/device is popular for architecture-specific instruction throughput modeling \citep{mendis2019ithemal, nemirovsky2017machine}, or for relating runtime to different input data for a given program using historical examples on the same platform \citep{huang2010predicting}. If a reference platform (or CPU simulator \citep{zheng2015learning}) is available, performance features from dynamic profiling can be used also to learn the relation to performance on another platform \citep{zheng2016accurate, zheng2017sampling, saeedmachine}.

\paragraph{Matrix Completion}

Modeling workload and platform characteristics jointly using a matrix completion approach such as matrix factorization can reduce the need for intrusive profiling or instrumentation. For example, Quasar and Paragon \cite{delimitrou2014quasar, delimitrou2013paragon} both use matrix factorization to predict QoS (quality-of-service) metrics for cloud orchestration, though neither use platform or application features. Other works have also explored matrix completion with side information for execution time prediction~\cite{pham2017predicting} using black-box methods.

\paragraph{Interference Modeling}

The difficulty of analytically modeling program interference has motivated several learning-based approaches, e.g., by constructing micro-benchmarks which are used as input features \cite{delimitrou2014quasar, delimitrou2013paragon} for predicting interference in the cloud. Previous edge-focused work also includes models targeting memory interference on embedded processors \citep{saeed2021learning}, OpenCL memory interference \citep{lee2017performance}, Simultaneous Multithreading contention \citep{moseley2005methods}, and virtual machine interference \citep{koh2007analysis}. These works train a device-specific model, while our method generalizes across heterogeneous devices, and is the first to integrate interference modeling into matrix factorization.

\paragraph{Uncertainty Quantification}

Previous works using uncertainty quantification for workload performance prediction rely on Bayesian Optimization, for example for cloud workload configurations \cite{alipourfard2017cherrypick} and time series workload demand forecasting \cite{zhou2022aquatope}. We instead base our approach on conformal prediction \cite{shafer2008tutorial, romano2019conformalized}, which is distribution-free and provides a mathematical guarantee of validity.

\begin{figure*}
    \centering
    \includegraphics[width=\textwidth]{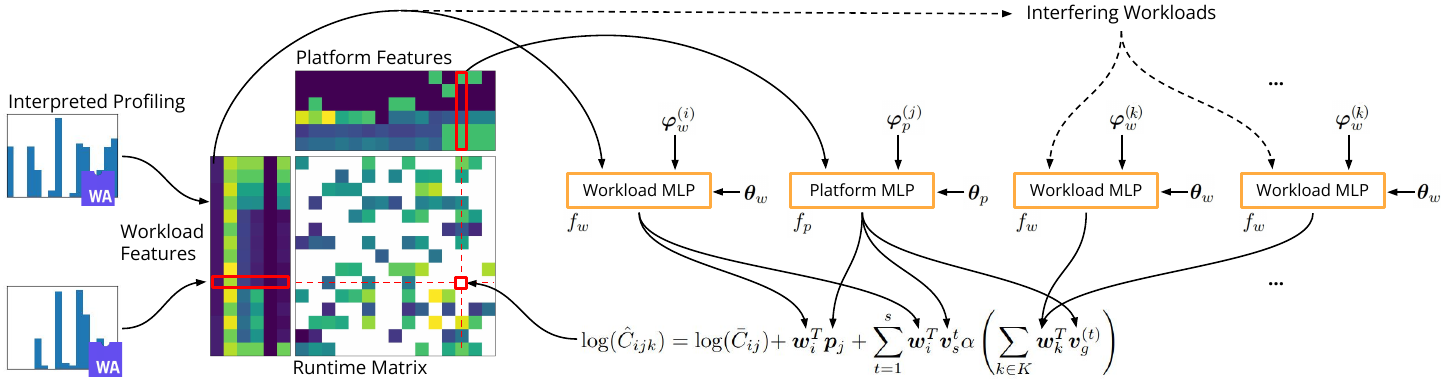}
    \vspace{-1.5em}
    \caption{Illustration of Pitot's interpreted profiling, matrix factorization, and interference model. Workload and platform embeddings $\bm{w}_i$, $\bm{p}_j$ are first computed by embedding networks $f_w, f_p$ from input features $\bm{x}_w^{(i)}, \bm{x}_p^{(j)}$ concatenated to learned features $\bm{\varphi}_w^{(i)}, \bm{\varphi}_p^{(j)}$. Then, for each (workload, platform) pair, Pitot adds the inner product $\bm{w}_i^T\bm{p}_i$ to the baseline $\bar{C}_{ij}$. If interfering modules are present, Pitot also computes the interference susceptibility $\bm{w}_i^T\bm{v}_s^t$ and magnitude $\bm{w}_k^T\bm{v}_g^{(t)}$ for each, and adds an interference term (Eq.~\ref{eq:multi_interference}). The resulting prediction is then compared to the observed runtime to train our model weights $\{\bm{\theta}_w, \bm{\theta}_p, \bm{\varphi}_w, \bm{\varphi}_p\}$.}
    \label{fig:model}
\end{figure*}

\section{Method}
\label{sec:method}

Figure~\ref{fig:model} illustrates our method, Pitot. As workloads are run on different platforms, possibly with other interfering workloads, we record the runtime of each (platform, workload, interference) tuple. We break down the key steps as follows:
\begin{enumerate}[(1), topsep=0pt, itemsep=0pt]
    \item Pitot uses a \textbf{log-residual objective}, which has a normalizing effect and allows us to handle heterogeneity across several orders of magnitude (Sec.~\ref{sec:log_residual}).
    \item A ``two-tower'' neural network-based \textbf{matrix factorization} model learns workload and platform embeddings from side information (Sec.~\ref{sec:matrix_completion_model}).
    \item Pitot adds a novel \textbf{interference prediction} term which models arbitrary interference patterns while also accounting for interference in the training data (Sec.~\ref{sec:interference_model}).
    \item If runtime bounds are required instead of a ``mean'' prediction, we \textit{conformalize} the outputs of our model (Sec.~\ref{sec:cqr}) to provide \textbf{uncertainty bounds}.
\end{enumerate}
Each step draws from insights across computer systems and machine learning literature and makes significant contributions to the efficacy of our method (Sec.~\ref{sec:ablations}), which together add up to over a 2x improvement over a naive application of existing methods (Sec.~\ref{sec:baselines}).

\subsection{Problem Formulation}
\label{sec:formulation}

\paragraph{Assumptions} Since predicting the runtime of unknown workloads consisting of arbitrary programs (and potentially infinite resource usage) is impossible in a general sense, we begin by making the following assumption: 
\begin{itemize}[topsep=0pt, itemsep=0pt]
    \item \textbf{Workloads are uniquely identifiable,} and if the nature of a workload changes, this can be identified externally. Thus, if a workload undergoes a phase shift, we assume that the shift can be identified, and the new phase treated as a new workload.
    \item \textbf{Each workload is observed at least once,} and we are never asked to predict the runtime of a workload using only features which are statically available.
    \item \textbf{Each platform is observed at least once,} and we are never asked to predict runtime on a platform using only a platform description.
    \item \textbf{Observations are exchangeable.} In other words, the order in which workloads are run does not matter, and the characteristics of platforms are constant over time.
\end{itemize}

\paragraph{Formulation} From these assumptions, we formulate our matrix completion problem as follows:
\begin{itemize}[topsep=0pt, itemsep=0pt]
    \item Workloads are indexed by $i = 1, 2, \ldots N_w$, and each have side information $\bm{x}_w^{(i)}$ (e.g. opcode count).
    \item Platforms are indexed by $j = 1, 2, \ldots N_p$, and are associated with side information $\bm{x}_p^{(j)}$ (e.g. CPU, runtime environment, memory information).
    \item We observe runtimes $C_{ijK}^*$ of workload $i$ running on platform $j$ for some subset $\mathcal{A}$ of (platform, workload, interference) tuples, where the interference $K$ is an arbitrary set of simultaneously running workloads.
\end{itemize}
We then try to predict runtimes $\hat{C}_{ijK}$ for unobserved $i,j,K$.

\subsection{Log-Residual Objective}
\label{sec:log_residual}

Execution time and other resource usage in computer systems are known to be extremely long-tailed \citep{tirmazi2020borg}, with the fastest and slowest programs, the fastest and slowest WebAssembly runtimes (ahead-of-time-compilers vs interpreters), and the strongest and weakest devices (modern x86 vs low-power embedded ARM) varying by several orders of magnitude in speed. When using an ``absolute'' objective such as $l_2$ error, this causes the total loss to be dominated by data with the largest magnitudes -- the slowest programs running on the slowest platforms -- where the same relative error accounts for disproportionate $l_2$ losses.

\paragraph{Log Runtime}

Using geometric instead of arithmetic averages is a well-established practice in benchmarking computer systems. This serves to prevent authors from arbitrarily weighting benchmarks by modifying each benchmark's total runtime to advantage methods of their choice\footnote{
    This practice has led to arithmetic averaging being labeled a ``benchmarking crime'' \cite{van2018benchmarking}, which we are careful not to commit.
}.

We find that this insight is not restricted to the fairness of benchmarking: modeling performance multiplicatively can also ``normalize'' the distribution of the data, which can be seen by applying the Central Limit Theorem (CLT). Modeling workloads as a collection of largely independent tasks, the CLT implies that the runtime should be well-behaved in distribution (and roughly normal). However, from a performance analysis perspective, while these tasks -- think a single instruction or system call -- may be largely independent \textit{conditioned on the platform}, they are highly correlated with the platform itself.

This implies that we should instead view workloads as a collection of performance characteristics which are \textit{multiplied} to find the runtime -- or characteristics which are added to find the log runtime, to which we can apply the CLT. As such, instead of simply normalizing workload runtimes by dividing by a metric \cite{delimitrou2013paragon}, we minimize the $l_2$ loss in \textit{log} space:
\begin{align}
\mathcal{L} = \sum_{(i, j, k) \in \mathcal{A}}||\log(C_{ijk}^*) - \log(\hat{C}_{ijk})||_2^2. \label{eq:log_loss}
\end{align}

\paragraph{Residual Objective}
\label{sec:residual_objective}
Under a log-objective, we can further normalize each workload and platform using a simple optimization procedure. Modeling workloads and platforms with a ``difficulty'' (i.e. instruction count) and ``speed'' (throughput), which, assuming linear performance scaling, are multiplied to find the runtime, we can learn a simple (interference-blind) model
\begin{align}
    \log(\bar{C}_{ij}) = \bar{w}_i + \bar{p}_j \label{eq:baseline}
\end{align}
for workload log ``difficulty'' $\bar{w}_i$ and platform log ``speed'' $\bar{p}_j$, which can be efficiently learned by alternating minimization (App.~\ref{appendix:linear_scaling}). Instead of directly predicting the runtime $\hat{C}_{ijk}$, we construct the rest of our model to predict the residual of the baseline $y_{ijk} = \log(C_{ijk}^*) - \log(\bar{C}_{ij})$.

This objective has the key advantage of being preserved under simple scaling. All else being equal, a job $i'$ consisting of $\gamma$ repetitions of job $i$ (i.e. $\hat{C}_{i'jk} = \gamma\hat{C}_{ijk}$ and $\bar{C}_{i'jk} = \bar{C}_{ijk}$) will satisfy $y_{ijk} = y_{i'jk}$. This provides significant advantages for prediction and interpretation, and can be seen by expanding $y_{i'jk}$:
\begin{equation}
\begin{aligned}
    y_{i'jk} &= \log(\gamma\hat{C}_{ijk}) - \log(\gamma\bar{C}_{ijk}) \\
    &= \log(\hat{C}_{ijk}) - \log(\bar{C}_{ijk})
    = y_{ijk}. \label{eq:residual-properties}
\end{aligned}
\end{equation}

\subsection{Matrix Factorization Model}
\label{sec:matrix_completion_model}

Matrix Factorization techniques decompose a partially observed low-rank target matrix into the product of multiple matrices. When decomposing a matrix into two matrices $\bm{C} = \bm{WP^T}$, each element $C_{ij}$ (the runtime of workload $i$ on platform $j$) is represented as the inner product $\bm{w}_i^T\bm{p}_j$ of the corresponding rows $\bm{w}_i$ in $\bm{W}$ and $\bm{p}_j$ in $\bm{P}$. This approach learns a common ``embedding space'': workloads and platforms with similar execution behavior should have similar $\bm{w}_i$ and $\bm{p}_j$ embeddings.

We predict the residual using matrix factorization (Fig.~\ref{fig:model}), with embeddings derived from the workload and module side information. Instead of analytical solutions such as \cite{chiang2015matrix}, we use the ``two-tower'' neural network architecture popular in recommender systems \citep{covington2016deep} to handle nonlinearity with respect to our side information $\bm{x}_w^{(i)}$ and $\bm{x}_p^{(j)}$. Specifically, the workload and platform embeddings $\bm{w}_i, \bm{p}_j \in \R^r$ for workload $i$ and platform $j$ are output by multi-layer perceptrons $f_w$ and $f_p$:
\begin{equation}
\begin{aligned}
    \bm{w}_i = f_w(\bm{x}_w^{(i)}, \bm{\varphi}_w^{(i)}; \bm{\theta}_w) \\
    \bm{p}_j = f_p(\bm{x}_p^{(j)}, \bm{\varphi}_p^{(j)}; \bm{\theta}_p).
\end{aligned}
\label{eq:mlp_embeddings}
\end{equation}
Here $f_w, f_p$ have weights $\bm{\theta}_w, \bm{\theta}_p$; $\bm{\varphi}_w^{(i)}, \bm{\varphi}_p^{(j)} \in \R^q$ are additional parameters\footnote{
This captures information that cannot be expressed as a function of the input features, for example if two workloads or platforms differ in hard-to-measure ways such as memory access patterns or memory latency, and is essential for the model to have sufficient representational capacity \ref{appendix:hyperparameters}.
} associated with each workload and platform that are appended to side information $\bm{x}_w^{(i)}$ and $\bm{x}_p^{(j)}$. The matrix factorization term is then added to Eq.~\ref{eq:baseline} to obtain the interference-blind prediction
\begin{align}
    \log(\hat{C}_{ij}) &= \log(\bar{C}_{ij}) + \bm{w}_i^T\bm{p}_j \label{eq:matrix_factorization}.
\end{align}

\paragraph{Model Architecture}

In our experiments, the workload and platform embedding networks $f_w, f_p$ each have 2 hidden layers of 128 units, and GELU activation on all hidden layers. When used with our dataset (Section~\ref{sec:dataset}), Pitot has 111,200 parameters split roughly equally between workload and platform embedding networks $\bm{\theta}_w, \bm{\theta}_p$, with a negligible number making up the additional learnable features $\bm{\varphi}_w^{(i)}, \bm{\varphi}_p^{(j)}$.

\subsection{Interference-Aware Prediction}
\label{sec:interference_model}

Inspired by the general principle of representation learning applied by matrix factorization -- learning a common embedding space for all workloads and platform, we add a term to our model that models interference in this embedding space. We begin by modeling the interference caused by a single workload using a low-rank ``interference matrix'' $\bm{F}_j$ for each platform $j$:
\begin{align}
    \log(\hat{C}_{ijk}) &= \log(\bar{C}_{ij}) + \bm{w}_i^T\bm{p}_j + \bm{w}_i^T\bm{F}_j\bm{w}_{k}. \label{eq:interference_aware}
\end{align}
This term, $\bm{w}_i^T\bm{F}_j\bm{w}_{k}$, models the log-performance penalty caused by workload $\bm{w_k}$ running alongside workload $\bm{w}_i$ on a platform with an interference matrix $\bm{F}_j$. Unlike previous works \cite{delimitrou2014quasar,delimitrou2013paragon}, this allows us to capture the key asymptotic benefits of matrix factorization: the ability to learn interactions between workloads and platforms simply by observing them, and without needing to explicitly model or benchmark these interactions.

\paragraph{Interpreting $\bm{F}_j$}

This interference term can be interpreted as an extension of matrix factorization that models interference in the common embedding space. Consider the SVD (singular value decomposition)-like decomposition
$\bm{F}_j = \sum_{t=1}^s \bm{v}_s^{(t)}\bm{v}_g^{(t)T}$
for a rank constraint $s$. Expanding our runtime prediction with this decomposition,
\begin{align}
    \bm{w}_i^T\bm{p}_j + \bm{w}_i^T\bm{F}_j\bm{w}_{k}
    &= \bm{w}_i^T\bm{p}_j + \sum_{t=1}^s \bm{w}_i^T\bm{v}_s^{(t)}\bm{w}_k^T\bm{v}_g^{(t)} \label{eq:susceptibility_magnitude} \\
    &= \bm{w}_i^T\left(\bm{p}_j + \sum_{t=1}^s \bm{v}_s^{(t)}\bm{w}_k^T\bm{v}_g^{(t)}\right) \label{eq:impact_magnitude}.
\end{align}
The model can be viewed (Eq.~\ref{eq:susceptibility_magnitude}) as the sum of $s$ types of interference, where the interfering module $\bm{w}_k$ causes a ``magnitude'' $\bm{w}_k^T\bm{v}_g^{(t)}$ of type $t$ interference, and the workload $\bm{w}_i$ has a ``susceptibility'' $\bm{w}_i^T\bm{v}_s^{(t)}$ to type $t$ interference. Factoring $\bm{w}_i$ out, this can also be interpreted (Eq.~\ref{eq:impact_magnitude}) as moving the platform embedding $\bm{p}_j$ in the direction $\bm{v}_s^{(t)}$ with magnitude $\bm{w}_k^T\bm{v}_g^{(t)}$ for each type $t$ of interference.

\paragraph{Multiple Interfering Workloads}

If multiple interfering workloads are present, we could extend our model by adding the interference magnitude $\bm{w}_k^T\bm{v}_g^{(t)}$ for each workload $k \in K$ using the susceptibility-magnitude representation (Eq.~\ref{eq:susceptibility_magnitude}). However, this assumes that workloads only cause multiplicative (i.e. log-additive) interference effects. We instead apply an activation function $\alpha$ to the total magnitude for each type:
\begin{align}
    \log(&\hat{C}_{ijk}) = \log(\bar{C}_{ij})  \nonumber \\
    &+ \bm{w}_i^T\bm{p}_j
    + \sum_{t=1}^s
        \bm{w}_i^T\bm{v}_s^{(t)}
        \alpha\left(\sum_{k \in K}\bm{w}_k^T\bm{v}_g^{(t)}\right).
    \label{eq:multi_interference}
\end{align}

This allows us to model the case that little interference is observed until a certain threshold, while also allowing any relationship between interference magnitude and runtime slowdown to be approximated with a sufficiently large $s$.

\paragraph{Model Architecture}

We learn $\bm{v}_s$ and $\bm{v}_g$ for each platform $j$ by adding additional output heads to $f_p$. We then compute our interference term using the factorization given in Eq.~\ref{eq:multi_interference}.

For the interference activation $\alpha$, we use a leaky ReLU activation with negative slope 0.1, since ordinary ReLU activation functions often lead to ``dead'' interference types (i.e. becoming extremely negative) due to poor initialization.

\subsection{Uncertainty Quantification using Conformalized Quantile Regression}
\label{sec:cqr}

When applying runtime prediction to resource allocation or system design, practitioners are often interested in not just the expected resource usage, but some (over-provisioned) bound which will be sufficient with high probability. Stated formally, for a workload distribution $C^*$, we would like to predict a regression output $\tilde{C}^{(\varepsilon)}$ such that
\begin{align}
    \Pr(C^* < \tilde{C}^{(\varepsilon)}) < \varepsilon \label{eq:conformal_regression}
\end{align}
for a one-sided \textit{target miscoverage rate} $\varepsilon > 0$ that represents the probability that the predicted runtime is insufficient for the target workload. Given the miscoverage rate as a constraint, our objective is to make our bound as ``tight as possible,'' minimizing the \textit{overprovisioning margin}
\begin{align}
    m = \ex[\max(\tilde{C}^{(\varepsilon)} - C^*, 0) / C^*], \label{eq:tightness}
\end{align}
which measures the relative resource overprovisioning that would occur using this prediction bound.
 
This motivates the usage of \textit{split conformal regression} \cite{shafer2008tutorial}, which uses a hold-out calibration set to provide statistically guaranteed bounds in probability (under the assumption of exchangeability). Split conformal regression can be used on any regression algorithm, and works by ``calibrating'' the regression output by adding a constant offset $\gamma$ to the output predictions $\hat{C}$:
\begin{align}
    \tilde{C}^{(\varepsilon)} = \hat{C} + \gamma: \quad \Pr(C^* < \hat{C} + \gamma) = \varepsilon.
\end{align}
However, when applied directly to a least-squares regression output, split conformal regression is not adaptive, and can only quantify the uncertainty of the model as a whole.

\paragraph{Conformalized Quantile Regression}

Quantile regression \cite{koenker1978regression} uses a ``pinball'' loss which estimates a \textit{target quantile} $\xi$ when minimized:
\begin{align}
    \mathcal{L} = \begin{cases}
        \xi(\hat{C} - C^*) & C^* > \hat{C} \\
        (1 - \xi)(\hat{C} - C^*) & C^* \leq \hat{C}
    \end{cases} \label{eq:pinball_loss}.
\end{align}
While this is not guaranteed to result in predictions which capture the target quantile, and in practice can be far from the target quantile, quantile regression does provide an \textit{adaptive} measure of uncertainty. By applying conformal regression to the output of quantile regression, we can obtain adaptive, yet calibrated, predictions; this is known as \textit{Conformalized Quantile Regression} (CQR) \cite{romano2019conformalized}.

\paragraph{Optimal Quantile Choice}

Common practice in CQR\footnote{This corresponds to $\xi = \varepsilon/2$ for the more commonly used two-sided CQR \cite{sousa2022improved, romano2019conformalized}.} is to set the target quantile as the miscalibration rate (i.e., $\xi = \varepsilon$); we refer to this as a ``naive'' application of CQR. No theoretical results justify this practice, and we find that in some cases, the quantile which results in the tightest bounds once calibrated  (i.e. minimizing Eq.~\ref{eq:tightness}) can vary greatly from the target miscoverage rate (App.~\ref{appendix:quantile_selection}).

Instead, we train Pitot on a spread of different $\xi$; then, for a target $\varepsilon$ at test time, we calibrate each output for $\varepsilon$. Finally, we calculate the tightness (Eq.~\ref{eq:tightness}) of each resulting predictor on the validation set, and select the best one. In addition to providing a modest boost to bound tightness, this also allows us to calibrate Pitot for an arbitrary $\varepsilon$ without needing to retrain the model, while only increasing the training time by $\approx5\%$.

\begin{figure*}[t]
    \centering
    \includegraphics[width=\textwidth]{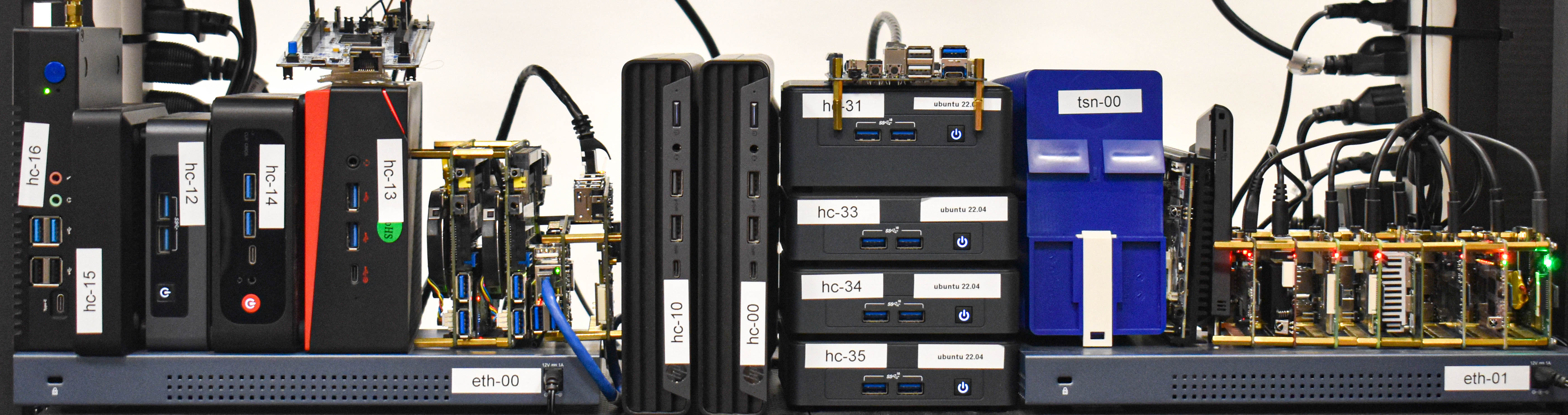}
    \vspace{-2.0em}
    \caption{Heterogeneous cluster test bench used to collect our dataset; our cluster includes Intel and AMD-based x86 computers, ARM A-class single board computers, as well as a RISC-V SBC and an ARM M-class microcontroller.}
    \label{fig:cluster}
\end{figure*}

\paragraph{Calibration Pools}

Conformalized quantile regression (and conformal prediction more broadly) only requires the assumption of \textit{exchangeability}: the ability to swap an observation in the calibration set with one in the test set without modifying the joint distribution of the calibration set. Notably, there is no requirement to have only \textit{one} calibration set: if sufficient data is available, we can partition the data based on some discrete random variable $I$ into multiple calibration and test set pairs while preserving exchangeability conditioned on $I$.

More homogeneous calibration sets are known to lead to smaller prediction intervals \cite{sousa2022improved}. Since we observe that program runtime is much more unpredictable when more interfering workloads are running, we split our data into different calibration pools depending on how many workloads were running simultaneously. As an added benefit, conditioning on the number of simultaneously-running workloads as $I$ allows Pitot to maintain \textit{conditional} exchangeability even under distribution shift of $I$.

\paragraph{Model Architecture}

Neural network-based quantile regression methods typically add multiple output heads, and train them jointly using a weighted objective. However, in embedding-based models such as matrix factorization, the output is a vector (or in our case, a collection of vectors), which would significantly increase training and inference time if naively duplicated for each target quantile $\xi$.

To limit this impact, we exploit the fact that only one of the embedding networks needs to have multiple outputs. Furthermore, Pitot's embeddings are not ``balanced'': the workload embedding consists of a single vector, while the platform embedding also must learn interference terms $\bm{v}_s$ and $\bm{v}_g$. As such, we learn multiple workload embeddings, and reuse the same platform embedding for each $\xi$.

\subsection{Training and Implementation}

Pitot is trained using stochastic gradient descent, and is extremely lightweight, with a single inference call taking $\approx$400Kflops, and training taking only 12.1 seconds (using a RTX 4090 on our dataset), including validation and checkpointing. For a detailed description of Pitot's hyperparameters, training procedure, and implementation, see Appendix~\ref{appendix:implementation}.

\section{Dataset}
\label{sec:dataset}

In order to develop and evaluate \textit{Pitot}, we assembled a heterogeneous compute cluster (Fig.~\ref{fig:cluster}) running a WebAssembly-based edge orchestration framework, and collected a dataset of 53,637 observations of different (workload, platform) pairs, and an additional 357,333 observations of workloads running on different platforms with up to 3 interfering workloads in the background. In total, this dataset represents approximately 80 hours of continuous data collection on our cluster. We provide a full description of our dataset and data collection methodology in Appendix~\ref{appendix:dataset}.

\paragraph{Workloads}
Our datatet uses 249 workloads drawn from the following benchmark suites:
\begin{itemize}[itemsep=0pt, topsep=0pt]
    \item Polybench \citep{pouchet_2015}: numerical floating-point-heavy kernels.
    \item MiBench \citep{guthaus2001mibench}: a diverse collection of miscellaneous benchmarks.
    \item UCSD Cortex Suite (including the San Diego Vision Benchmark Suite) \citep{thomas2014cortexsuite}: computer vision and machine learning benchmarks.
    \item Libsodium benchmark suite: cryptography benchmarks from the Libsodium test and benchmark suite.
    \item Python: 12 benchmarks written for CPython, run on a WebAssembly-compiled CPython using WASI.
\end{itemize}

While some benchmarks required light modifications to compile to WebAssembly and run on our platforms, continual efforts to improve software support such as extensions to the WebAssembly Standard Interface or by exposing linux system calls \cite{wasmwali} raise the possibility of bringing more legacy applications to WebAssembly in the future.

\paragraph{Platforms}
We assembled a cluster (Fig.~\ref{fig:cluster}) of 24 different devices ranging from a microcontroller\footnote{Notably, the microcontroller executes some of the smallest benchmarks faster than many linux-based platforms due to the absence of operating system overhead.} and various single-board computers to x86 desktop computers, and ran each benchmark using a suite of different WebAssembly runtimes on each device (App.~\ref{appendix:cluster}).

\paragraph{Input Features}
Bytecode-based virtualization allows a system to ``inspect'' program execution in a cross-platform way instead of treating programs as ``black box'' binaries. To take advantage of this, we count the number of times each opcode was executed using an instrumented interpreter, which we collect as workload features $\bm{x}_w$ (similar to \cite{kuperberg2007predicting}). We also collect data about each platform such as CPU and WebAssembly host information, which we provide as platform features $\bm{x}_p$. For a detailed specification of these features, see Appendix \ref{appendix:side_information}.

\paragraph{Limitations}
Our dataset is limited to a set of publicly available benchmarks that we were able to run on our test cluster; these benchmarks are primarily compute-bound, though some do feature a significant amount of filesystem I/O. In our dataset and formulation, we also assume that workloads either have a constant input data distribution, or perform computations that do not depend on the input data (e.g., control algorithms); workloads with phase changes due to alternate input data (e.g., data analysis) are treated as separate workloads similar to other works \cite{zheng2015learning, delimitrou2014quasar, delimitrou2013paragon}. As such, different data inputs or data sizes for the same binary or source code in our benchmarks (e.g., different python scripts for the same Python binary) are treated as separate workloads.

\begin{figure*}[t]
\begin{subfigure}[t]{0.5\textwidth}
    \centering
    \includegraphics[width=\columnwidth]{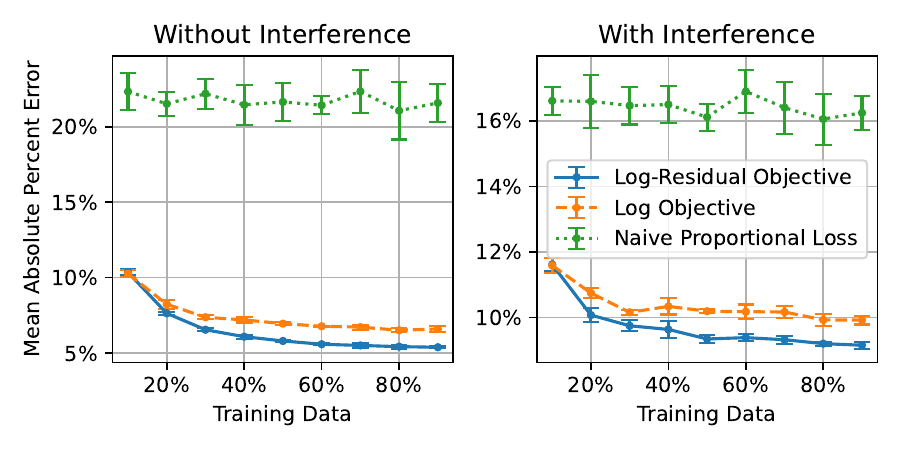}
    \vspace{-2em}
    \caption{Different loss formulations}
    \label{fig:ablation_objective}
\end{subfigure}\begin{subfigure}[t]{0.5\textwidth}
    \centering
    \includegraphics[width=\columnwidth]{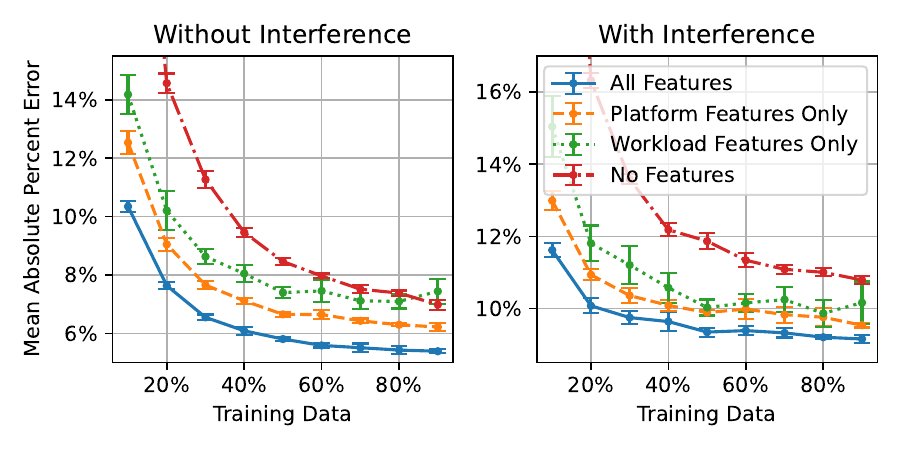}
    \vspace{-2em}
    \caption{Inclusion of workload and platform features}
    \label{fig:ablation_features}
\end{subfigure}

\begin{subfigure}[t]{0.5\textwidth}
    \centering
    \includegraphics[width=\columnwidth]{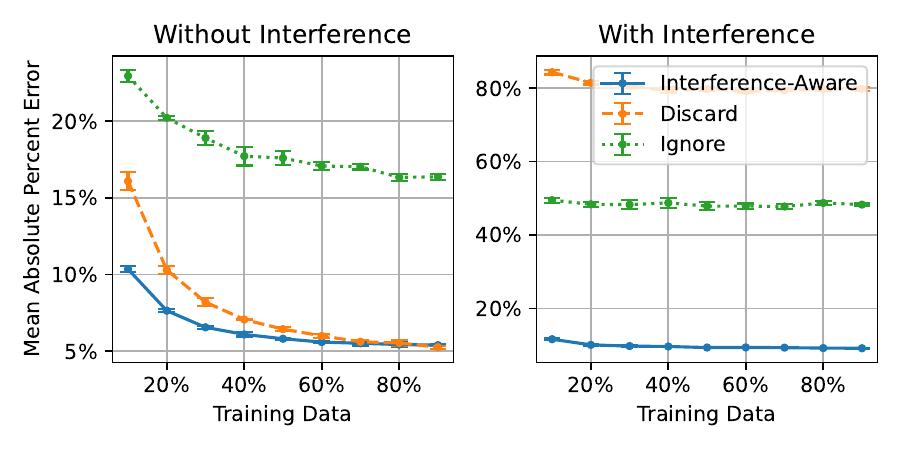}
    \vspace{-2em}
    \caption{Interference handling}
    \label{fig:ablation_interference}
\end{subfigure}\begin{subfigure}[t]{0.5\textwidth}
    \centering
    \includegraphics[width=\columnwidth]{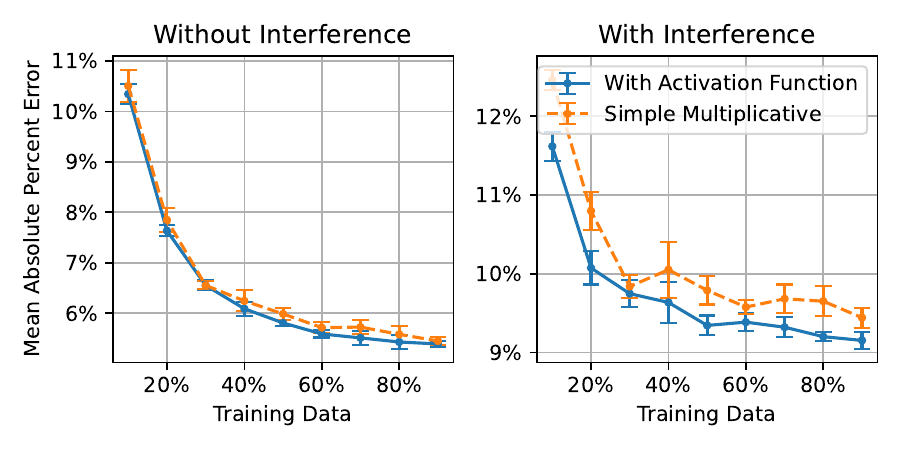}
    \vspace{-2em}
    \caption{Activation function for multiple interfering workloads}
    \label{fig:ablation_relu}
\end{subfigure}
\caption{Ablations of key aspects of our Pitot. Each figure shows the mean absolute percent error ($\pm 2$ standard errors) for varying amounts of training data; error for test data with and without interference are shown separately.}
\label{fig:ablations}
\end{figure*}

\section{Results}
\label{sec:results}

To evaluate our contributions, we ran ablations showing how each aspect of Pitot contributes to its superior performance (Sec.~\ref{sec:ablations}) compared to existing approaches (Sec.~\ref{sec:baselines}). We also show how Pitot learns interpretable embeddings that can potentially be used for other tasks (Sec.~\ref{sec:interpretation}).

\subsection{Experiment Setup}
\label{sec:prediction_experiments}

\paragraph{Evaluation}

To measure data efficiency, we evaluated each method in training data splits of $10\%, 20\%, \ldots 90\%$ of the data. We trained 5 replicates for each configuration (method and train split size), with each replicate independently receiving a train and test set. Within the training set, 80\% was used for actual method training, while the remaining 20\% was used for validation and calibration (where applicable).

Since predicting the runtime of workloads running with interference is intrinsically harder than for workloads running in isolation (and the weighting between the two objectives is arbitrary), we also show the performance on test data with and without interfering workloads separately.

\paragraph{Error}

To evaluate the accuracy of our average runtime predictions, we report the Mean Absolute Percent Error (MAPE) on the holdout test set $\mathcal{A}^c$.

\paragraph{Tightness}

To evaluate the tightness of our predicted uncertainty bounds, we compute the overprovisioning margin (Eq.~\ref{eq:tightness}) for miscoverage rates $\varepsilon = 0.1, 0.09, \ldots 0.01$, which measures the average excess of the predicted bound compared to the actual runtime.

Note that while models trained to predict the expected runtime (i.e. squared loss minimization) can be calibrated and used to predict (conformalized) bounds, it is not appropriate to evaluate models trained to predict a target quantile using an error metric. As such, we evaluate error on a version of Pitot trained with a squared loss (Eq.~\ref{eq:log_loss}) and evaluate tightness on a version trained with quantile regression (Eq.~\ref{eq:pinball_loss}).

\subsection{Method Ablations}
\label{sec:ablations}

Pitot draws its accurate predictions and tight bounds not from a single key insight, but from the collection of improvements we highlight in this paper. To demonstrate this, we ran ablations on the contribution of each improvement.

\paragraph{Hyperparameter Ablations}

We provide ablations for key hyperparameters of Pitot in Appendix~\ref{appendix:hyperparameters}; as long as sufficient representational capacity is provided, Pitot is not sensitive to the choice of hyperparameters. 

\paragraph{Log-residual Objective}

Figure~\ref{fig:ablation_objective} shows the impact of our log-residual objective compared to an ordinary log objective and a naive proportional loss on the prediction error. Due to the large variation in runtime magnitudes, a naive proportional loss cannot achieve reasonable error. Adding a residual objective also significantly improves accuracy, especially as more of the dataset is observed.

\paragraph{Side Information}

Adding side information in the form of workload and platform features can significantly boost accuracy, especially when only a small amount of data is observed. In our experiments, both platform and workload features significantly decrease error, especially when used together (Fig.~\ref{fig:ablation_features}). Platform features do have a significantly higher marginal impact than workload features; this is likely due to the presence of similar platforms in our dataset (e.g. different Cortex-A53 processors) which the model can pick up on. In larger datasets where many similar workloads are present (e.g. different versions of the same workload), this relationship would likely be reversed.

\paragraph{Interference Prediction}

We compare our interference-aware method (Eq.~\ref{eq:interference_aware}) to two alternate strategies: \textit{discard}, where we discard any observations with interfering workloads, and \textit{ignore}, where all observations are treated the same, regardless of any interference.

Discarding data leads to a low error floor when predicting data without interference, though the predictor cannot capture interference. On the other hand, ignoring the effects of interference leads to a predictor with much higher error, since the effects of interference end up ``averaged in'' to all predictions. By modeling interference, Pitot can also use observations with interference to improve overall accuracy: when less data is observed, the interference-aware model has a significantly lower error than an interference-blind model only trained on ``clean'' data (Fig.~\ref{fig:ablation_interference}).

\paragraph{Multiple Interfering Workloads}

In order to model interference relationships beyond a simple multiplicative model, Pitot includes an activation function applied to the learned interference ``magnitude'' (Eq.~\ref{eq:multi_interference}). Figure~\ref{fig:ablation_relu} shows the modest but significant impact of including this activation function, indicating that ``interference thresholds'' have a small but significant effect.

\paragraph{Uncertainty Quantification}

\begin{figure}
    \centering
    \includegraphics[width=\columnwidth]{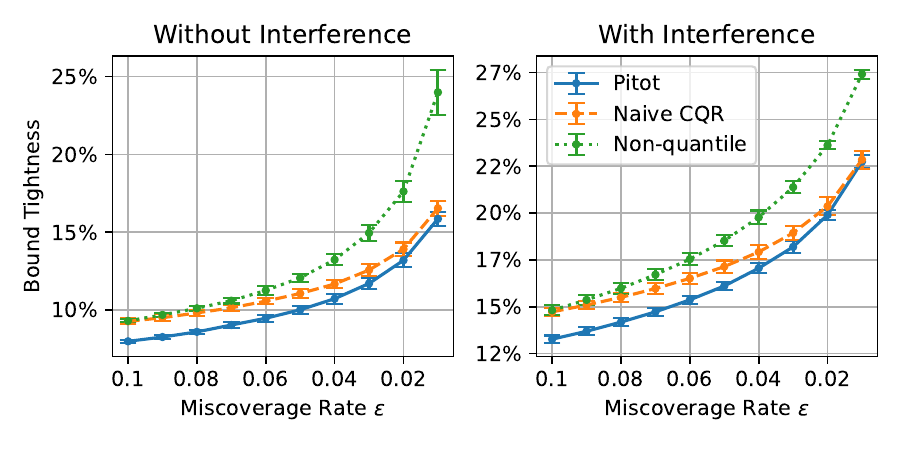}
    \vspace{-2.5em}
    \caption{Bound tightness ($\pm 2$ standard errors) of our conformalized quantile regression algorithm compared to naive approaches for varying miscoverage rates when trained on 50\% of the dataset.}
    \label{fig:ablation_cqr}
\end{figure}

Figure~\ref{fig:ablation_cqr} shows the impact of our CQR-based uncertainty quantification method compared to a naive application of CQR and calibrating an ordinary predictor not trained using quantile regression. CQR produces significantly tighter bounds especially for small miscoverage rates, while our quantile choice method improves CQR significantly at larger miscoverage rates.

\subsection{Baselines}
\label{sec:baselines}

\begin{figure*}[t]
\centering
\begin{subfigure}[t]{0.49\textwidth}
    \centering
    \includegraphics[width=\columnwidth]{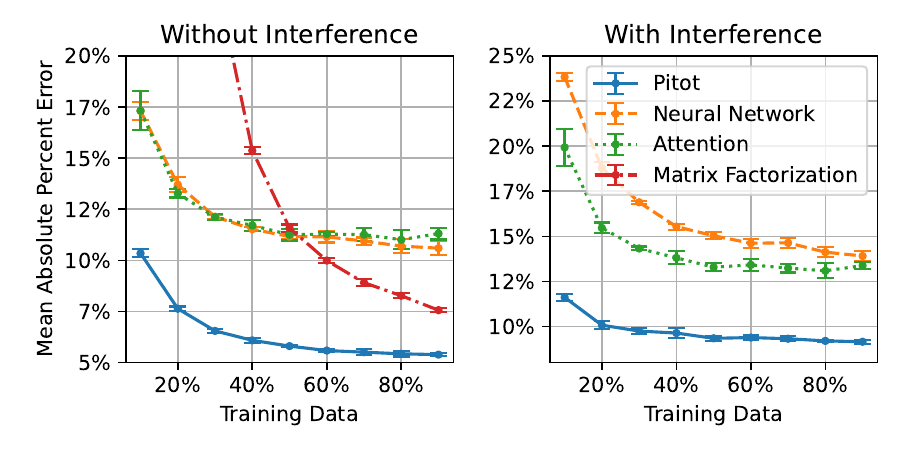}
    \vspace{-2em}
    \caption{Error for different amounts of training data. Matrix Factorization is not visible due to exceeding 75\% error in all cases; see App.~\ref{appendix:uncropped} for an uncropped version.}
    \label{fig:baseline_error}
\end{subfigure}
\hfill
\begin{subfigure}[t]{0.49\textwidth}
    \centering
    \includegraphics[width=\columnwidth]{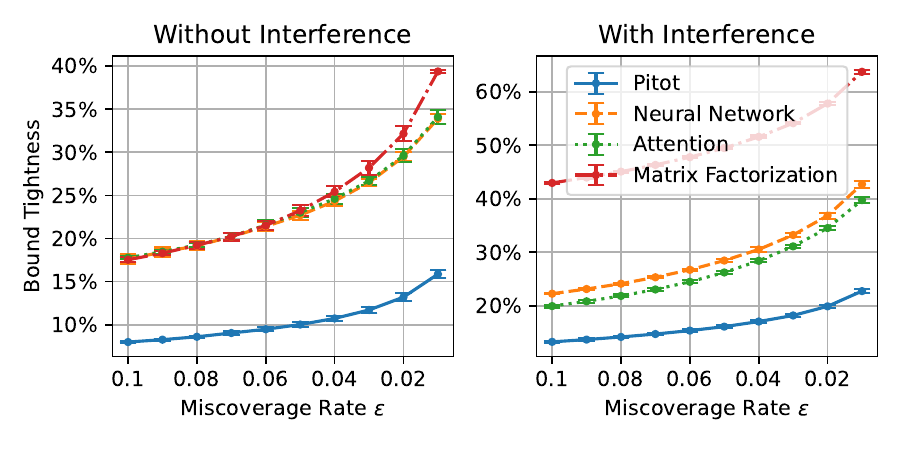}
    \vspace{-2em}
    \caption{Bound tightness for varying miscoverage rates when trained on 50\% of data; see Appendix~\ref{appendix:baseline_width_full} for other data splits.}
    \label{fig:baseline_width}
\end{subfigure}
\caption{Comparisons against baselines; Pitot has significantly less error and tighter bounds in all settings that we evaluated. \textit{Matrix factorization} performs better relative to other baselines when more data is observed, but lacks data efficiency. Also, while \textit{attention} is less accurate than our interference model, it does perform better than \textit{neural network}, supporting the need for better interference modeling.}
\label{fig:baseline}
\end{figure*}

While no prior works tackle runtime prediction in the same setting (interference-aware runtime prediction with interference in a heterogeneous environment), we compare Pitot with several baselines designed from elements drawn from state-of-the-art prediction algorithms. For additional details on our baselines, see Appendix~\ref{appendix:baselines}.

\paragraph{Matrix Factorization} 

As our first baseline, we use a matrix factorization model similar to \cite{delimitrou2014quasar, delimitrou2013paragon} which predicts the (log) runtime by learning a feature vector for each workload and platform. This model does use workload or platform features, and is not interference-aware (and discards any observations with interference) since matrix factorization cannot be easily extended to explicitly predict the impact of interference on runtime\footnote{While (platform, workload, interference) tensor completion is possible, the size increases exponentially with each additional interfering workload, quickly leading to unworkable sparsity. While \cite{delimitrou2013paragon} does include a matrix factorization-based interference model, it relies on a microbenchmarking approach for cloud applications, and cannot predict the runtime time impact of an arbitrary interfering workload --- only a relative measure of interference impact and susceptibility.}.

\paragraph{Neural Network}

For a stronger baseline, we use a neural network-based model which takes our workload and platform features as inputs similar to \cite{pham2017predicting} to generate a base, interference-blind predictions. To handle interference, we add a second neural network which predicts an interference multiplier \cite{saeed2021learning} for each interfering workload.

\paragraph{Attention}

For our final baseline, we replace the simple multiplicative interference model in our neural network baseline with an attention mechanism followed by an output head which predicts a single interference multiplier which is applied to the base prediction. This attention mechanism is somewhat similar to Pitot's interference model, which can be thought of as a simple attention mechanism with a theory-informed output function instead of a neural network.

\paragraph{Comparison with Baselines}

Pitot has both significantly better accuracy (Fig.~\ref{fig:baseline_error}) and tighter bounds (Fig.~\ref{fig:baseline_width}) compared to each baselines for all evaluation settings (App.~\ref{appendix:baseline_width_full}). Unlike the ``pure'' \textit{matrix factorization} baseline without workload or platform features, Pitot is highly data efficient, and has lower error and tight bounds even when only a small amount of data is observed. Calibrating the \textit{neural network} and \textit{attention} baselines using split conformal regression to predict uncertainty bounds, Pitot also has much better accuracy and tighter bounds, demonstrating the advantage of our principled approach compared to generic neural networks. 

Overall, Pitot is an extremely accurate predictor which can have up to 48\% less error and 58\% tighter bounds (and an average of 36\% error and 44\% tighter) compared to the next best baseline across each evaluation setting.

\subsection{Model Interpretation}
\label{sec:interpretation}

\begin{figure}
    \centering
    \includegraphics[width=\columnwidth]{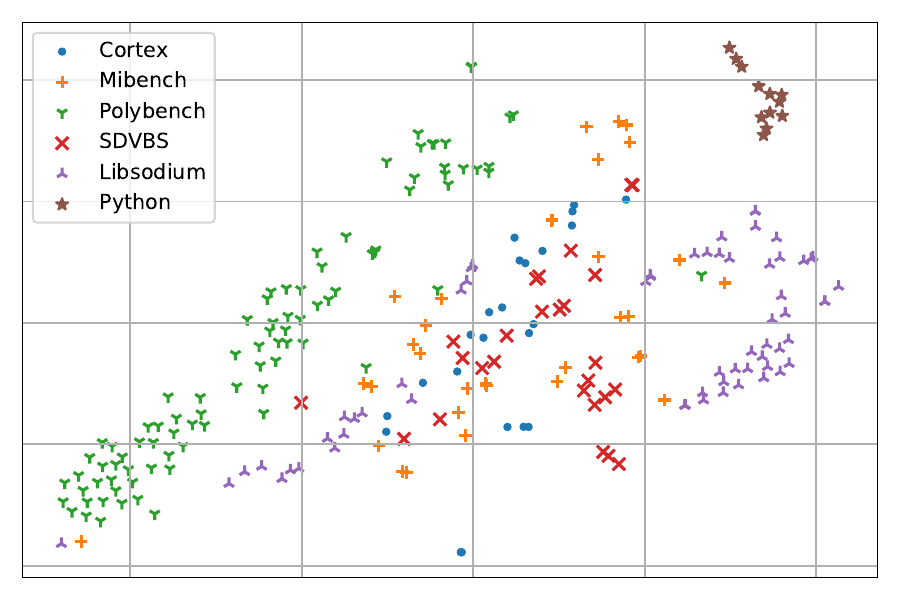}
    \vspace{-2.5em}
    \caption{2-dimensional t-SNE of workload embeddings by benchmark suite; the axes shown do not haves any particular meaning.}
    \label{fig:tsne_workload_preview}
\end{figure}

Unlike black-box methods, Pitot's matrix factorization-based approach learns embeddings $\bm{w}_i, \bm{p}_j$ for each workload and platform. These embeddings map similar workloads and platforms to nearby points in the embedding space, thereby quantifying their performance characteristics, and could be used for downstream tasks such as clustering or anomaly detection. Figure~\ref{fig:tsne_workload_preview} shows a t-distributed Stochastic Neighbor Embedding (t-SNE) of our learned workload embeddings. We can observe a clear clustering of workloads by benchmark suite\footnote{Especially for relatively homogenous benchmark suites such as Polybench and Libsodium.}, demonstrating the interpretability of the platform embeddings. For additional visualizations of platform and interference embeddings, see Appendix~\ref{appendix:visualizations}.

\section{Conclusion}

We formulate workload runtime prediction as an interference-aware matrix completion problem, and present our solution, Pitot, which combines several novel contributions including a novel ``log-residual'' training objective, interference-aware matrix factorization, and uncertainty quantification using conformalized quantile regression to make predictions which are far more accurate and tightly bounded than combining existing methods.

Possible applications of Pitot such as hardware-software co-design and edge orchestration could also benefit from future extensions that build on our work, including efficient online learning or statistical bounding techniques for miscoverage rates beyond what is possible using distribution-free approaches such as conformal prediction. We hope that our dataset -- which we intend to be a ``living'' project that will be updated as we add new platforms to our cluster, new benchmarks to our collection, and new measurement types -- will enable other to tackle these problems.

\paragraph{Limitations} While our dataset is substantial, it cannot fully represent distributed systems deployed at scale. Edge orchestration frameworks for cyber-physical systems \cite{silverlinewasmcon} have not yet seen widespread deployment, leading to a lack of real-world applications which can be used to test performance prediction algorithms; as such, Pitot is a methodological proof of concept, not a trained and deployable model. However, with sufficient buy-in from a large organization or deployment framework, we believe that our method could be applied at scale and provide substantial utility to practitioners and system managers alike.

Machine learning workloads present another challenge: although predicting the performance of machine workloads is of great relevance in edge computing, the lack of software portability and standardized workload/platform interfaces makes cross-platform runtime analysis impractical. Thus, as the edge-ML ecosystem matures, it may enable future work to bring Pitot -- or a similar technique -- to machine learning workloads.

\newpage
\bibliographystyle{mlsys2025}
\bibliography{ref}

\clearpage
\appendix

\section{Glossary of Notation}

We provide a glossary of the symbols used in this paper for convenience in Table~\ref{tab:glossary}.

\begin{table*}
    \caption{Glossary of notation used in this paper.}
    \vspace{0.2em}
    \centering
    {\small
    \begin{tabular}{c lp{4in} }
    \toprule
    Symbol & Description & Notes \\
    \toprule
    $N_w, N_p$ & Unique workloads, platforms & $N_w = 249, N_p = 231$ in our dataset. \\
    $i, j$ & Workload, platform index & $1 \leq i \leq N_w$ and $1 \leq j \leq N_p$. \\
    $k$ & Set of interfering workloads & $\forall l \in k: 1 \leq l \leq N_w$. We sometimes abuse notation and use $k$ as an index when $k$ is a singleton. \\
    $\mathcal{A}$ & Dataset & Contains all observed (workload, platform, interference) tuples. \\
    $C^*_{ijk}$ & Actual runtime & \\
    $\hat{C}_{ijk}$ & Predicted runtime & Non-interference-aware predictions are abbreviated $\hat{C}_{ij}$. \\
    $\bar{C}_{ij}$ & Baseline prediction & Linear scaling baseline predicted runtime. \\
    $\bar{w}_i, \bar{p}_j$ & Baseline parameters & Log workload ``difficulty'' and platform ``speed'' \\
    $\bm{x}_w$ & Workload features & Log opcode counts. \\
    $\bm{x}_p$ & Platform features & CPU architecture, WebAssembly runtime information. \\
    $\bm{w}_i, \bm{p}_j$ & Learned embeddings & Dimensionality $r=128$ workload, platform embeddings. \\
    $f_w, f_p$ & Embedding networks & Generates workload and platform embeddings \\
    $\bm{\theta}_w, \bm{\theta}_p$ & Embedding network weights & \\
    $\bm{\varphi}_w^{(i)}, \bm{\varphi}_p^{(j)}$ & Extra learned features & Has dimension $q=1$. \\
    $\bm{F}_j$ & Interference matrix & We never \textit{explicitly} compute $\bm{F}_j$. \\
    $\bm{v}_s^{(t)}$ & Interference susceptibility & Associated with a platform $j$ and interference type $t$ ($1 \leq t \leq s=2$). \\
    $\bm{v}_g^{(t)}$ & Interference magnitude & Associated with a platform $j$ and interference type $t$ ($1 \leq t \leq s=2$). \\
    $\alpha$ & Interference activation & Activation function (Leaky ReLU) for multiple interfering workloads \\
    $\varepsilon$ & Target miscoverage rate & For conformal regression. \\
    $\xi$ & Target quantile & For quantile regression. \\
    \toprule
    \end{tabular}
    }
    \vspace{-1em}
    \label{tab:glossary}
\end{table*}

\section{Additional Method Details}

In this section, we provide details regarding the architecture, training, calibration, and implementation of Pitot. Our code and dataset are also open source, and can be found at \url{https://github.com/wiseLabCMU/pitot}; an archival copy is also available at
\url{https://zenodo.org/records/14977004}.

\subsection{Linear Scaling Baseline}
\label{appendix:linear_scaling}

In this section, we provide a brief sketch of how we learn the parameters of the baseline model. We refer to this model as the \textit{Linear Scaling Baseline} since it corresponds to common benchmarking practice where (geometric) mean benchmarking scores are used to estimate a linear relationship between platforms.
Note that the linear scaling baseline is only learned from data collected with no interfering workloads running in the background.

\begin{proposition}
Since the log-loss (Eq.~\ref{eq:log_loss}) is convex for $\bar{m}_i$ and $\bar{p}_j$, we can efficiently learn the linear scaling model $\log(\bar{C}_{ij}) = \bar{w}_i + \bar{p}_j$ from $C_{ij}^*$ by alternating minimization over $\bar{w}_i$ and $\bar{p}_j$ using the update rule
\begin{align}
    \bar{m}_i = \frac{\sum_{i, j\in \mathcal{A}} \log(C_{ij}^*) - \bar{p}_j}{\sum_{i, j\in \mathcal{A}} 1}, \label{eq:baseline_update_rule}
\end{align}
with a similar rule applying for $\bar{p}_j$.
\end{proposition}

Convexity can easily be verified by noting that the loss (Eq.~\ref{eq:log_loss}) is the sum of convex quadratics (with respect to $\bar{m_i}$ and $\bar{p_j}$ individually). The update rule (Eq.~\ref{eq:baseline_update_rule}) then follows by differentiating and solving for $\partial \mathcal{L}/\partial \bar{m}_i = 0$, with the update rule for $\bar{p}_j$ being symmetric to $\bar{m}_i$.

\subsection{Quantile Selection}
\label{appendix:quantile_selection}

\begin{figure}
    \centering
    \includegraphics[width=\columnwidth]{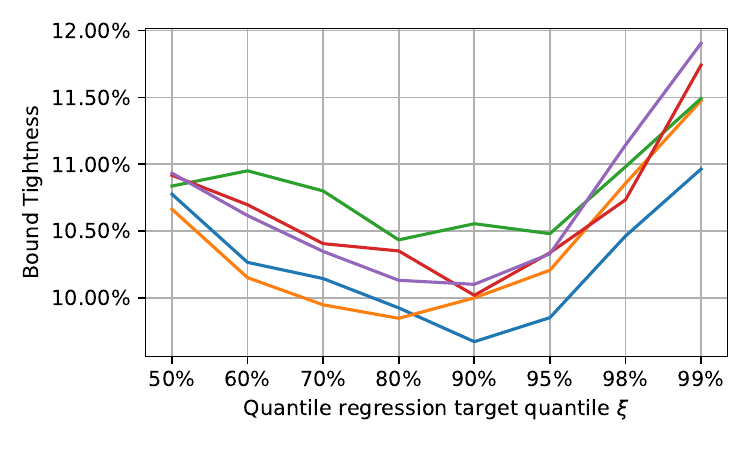}
    \vspace{-2.5em}
    \caption{Bound tightness (overprovisioning margin) resulting from different quantile regression target quantiles $\xi$ for 5 different replicates. The optimal target quantile is between 80\% and 90\%, compared to the calibration target miscoverage ratio of 95\%.}
    \label{fig:optimal_quantile}
\end{figure}

In (one-sided) conformalized quantile regression, using the same target quantile as the desired miscoverage ratio (i.e. $\xi = \varepsilon$) can be significantly less than optimal. Figure~\ref{fig:optimal_quantile} shows an illustrative example with replicates trained on 50\% of the dataset for prediction without interference, with a miscoverage ratio of $\varepsilon=0.05$. For each replicate, the optimal quantile regression target quantile which results in the narrowest overprovisioning margin after calibration is between 80\% and 90\%.

In our experiments, we also observe that small changes in $\xi$ have a larger impact on the resulting overprovisioning margin closer to $\xi = 100\%$. As such, we train target quantiles of \{50\%, 60\%, 70\%, 80\%, 90\%, 95\%, 98\%, 99\%\}, with more target quantiles close to 100\%.

\subsection{Model Training}
\label{appendix:implementation}

\paragraph{Multi-objective Optimization}

Pitot uses several different optimization objectives:
\begin{itemize}[topsep=0pt, itemsep=0pt]
    \item Interference mode: in order to balance the influence of prediction with and without interference and better utilize GPU acceleration (Appendix~\ref{appendix:implementation}), each interference mode (without interference and with 2, 3, and 4 simultaneously running workloads) is treated as a different objective.
    \item Quantile regression: for each interference mode, each target quantile is also a different optimization objective (Section~\ref{sec:cqr}).
\end{itemize}

In order to define a single optimization objective for gradient descent, we assign a weight to each objective:
\begin{itemize}[topsep=0pt, itemsep=0pt]
    \item To account for the increased difficulty and randomness of interference, and thus the ``higher quality'' of data collected without interfering workloads, we assign a higher weight to prediction without interference (Appendix~\ref{appendix:hyperparameters}).
    \item Each quantile regression output is given equal weight.
\end{itemize}

\paragraph{Training Details}

Pitot (and all of our baselines) were trained using the AdaMax optimizer (i.e. the $l_\infty$ variant of Adam) with default hyperparameters (learning rate = 0.001, $\beta_1=0.9, \beta_2=0.999$) and a batch size of 2048 (split equally across non-interference, 2, 3, and 4-way interference objectives).

Each model was trained for 20,000 steps, which we found was enough for convergence in all cases. During training, we evaluated each model every 200 steps, and returned the checkpoint which had the lowest validation loss for testing.

\paragraph{Implementation}

Our algorithm is implemented in JAX \cite{jax2018github}. While our dataset contains many ($N=410970$) data points, each data point uses a small amount of memory, consisting only of platform, workload, and interfering workload indices, which point to shared platform ($N_p=231$) and workload ($N_w=249$) features. As such, we make a number of optimizations in our implementation which target this data regime:
\begin{itemize}[topsep=0pt, itemsep=0pt]
    \item All data is stored in GPU memory at all times.
    \item Since our batch size (2048) is relatively large compared to our matrix (231 platforms, 249 workloads), we always compute all module and device embeddings $\bm{w}_i$ and $\bm{p}_j$, and index the ones that we need.
\end{itemize}

As an additional optimization, when training on data with interference, each additional source of interference adds additional nodes to the compute graph that are only used when interference is present. As such, we separately sample fixed-sized batches of 512 samples from each degree of interference instead of randomly drawing a batch of 2048 data points from the entire dataset at once in order to maximize GPU parallelism (i.e. allowing all operations to have a fixed dimension across batches) while avoiding wasted compute (i.e. if the results of unused computations are ignored).

With these optimizations, our method is very cheap to train, and has a median per-replicate training time of 11.5 seconds (or 12.1 seconds for the multi-objective quantile regression version) on a RTX 4090 GPU across 45 different runs.

\subsection{Baseline Details}
\label{appendix:baselines}

\paragraph{Common settings} To make our baselines more competitive, each baseline was also trained to predict runtime in the log domain. The baselines were also trained in the same way as Pitot (20,000 steps with batch size 2048, etc).

\paragraph{Matrix Factorization} Our matrix factorization baseline uses the same number of features ($r=32$) as we found to be optimal for Pitot, and can be thought of as Pitot without our log-residual objective, workload or platform features, interference modeling, and uncertainty quantification method.

\paragraph{Neural Network} The neural network baseline uses two neural networks with two hidden layers of 256 units and the GELU activation (twice as large as Pitot):
\begin{enumerate}[(1), topsep=0pt, itemsep=0pt]
    \item The ``base'' network concatenates the workload and platform features of each data point as input, and predicts a single interference-blind runtime which is used on workloads running in isolation.
    \item The ``interference'' network concatenates two sets of workload features (current workload and interfering workload) and one set of platform features as an input, and predicts an interference multiplier.
\end{enumerate}

The interference network computes an interference multiplier for each interfering workload; the base prediction is multiplied by each interference multiplier to generate the final interference-aware runtime prediction.

\begin{table*}[]
    \caption{Cluster devices with the CPU vendor, model, and microarchitecture.}
    \vspace{0.5em}
    \centering
    {
    \small
    \begin{tabular}{cccc}
        \toprule
        Model & \multicolumn{2}{c}{CPU} & Architecture \\
        \toprule
        NUC 8 & Intel & i7-8650U & Skylake \\
        NUC 4 & Intel & i3-4010U & Haswell \\
        Generic ITX & Intel & i7-4770TE & Haswell \\
        Compute Stick & Intel & x5-Z8330 & Silvermont \\
        NUC 11 & Intel & i5-1145G7 & Tiger Lake \\
        NUC 11 & Intel & i7-1165G7 & Tiger Lake \\
        Mini PC & Intel & N4020 & Goldmont Plus \\
        EliteDesk 805 G8 & AMD & R5-5650G & Zen 3 \\
        Mini PC & AMD & R5-4500U & Zen 2 \\
        Mini PC & AMD & R3-3200U & Zen 1 \\
        Mini PC & AMD & A6-1450 & Jaguar \\
        \toprule
    \end{tabular}
    \hspace{0.2cm}
    \begin{tabular}{cccc}
        \toprule
        Model & \multicolumn{2}{c}{CPU} & Architecture \\
        \toprule
        RPi 4 Rev 1.2 & Broadcom & BCM2711 & Cortex-A72 \\
        RPi 3B+ Rev 1.3 & Broadcom & BCM2837B0 & Cortex-A53 \\
        Banana Pi M5 & Amlogic & S905X3 & Cortex-A55 \\
        Le Potato & Amlogic & S905X & Cortex-A53 \\
        Odroid C4 & Amlogic & S905X3 & Cortex-A55 \\
        RockPro64 & RockChip & RK3399 & Cortex-A72 \\
        Rock Pi 4b & RockChip & RK3399 & Cortex-A72 \\
        Renegade & RockChip & RK3328 & Cortex-A53 \\
        Orange Pi 3 & Allwinner & H6 & Cortex-A53 \\
        Starfive VF2 & SiFive & U74 & RISC-V \\
        Nucleo-F767ZI & STMicro & STM23F767ZI & Cortex-M7 \\
        \toprule
    \end{tabular}
    }
    \vspace{-1em}
    \label{tab:cluster}
\end{table*}

\paragraph{Attention} The attention network uses a (single-headed) attention mechanism to predict the interference generated by a set of interfering workloads instead of assuming a simple multiplicative relationship between pairs of workloads. Like the neural network baseline, a neural network with two hidden layers of 256 units and GELU activation is used to generate a ``base'' prediction. To add an attention mechanism, this network also generates a query vector.

To model interference, a second embedding network (also with two hidden layers of 256 units and GELU activation) generates key and value vectors. The query vector is used to index the weight of the value vector across each interfering workload according $\langle \text{key}, \text{query}\rangle$ product, and an output network with a single hidden layer produces the final interference multiplier. We tuned the key/query vector dimension and output network hidden layer size, arriving at a vector dimension of 8 and an output hidden layer of 32.

\section{Dataset}
\label{appendix:dataset}

Using our heterogeneous cluster (Figure~\ref{fig:cluster}), we collected a large dataset which includes a range of different workloads, compute platforms, and varying levels of interference. In this section, we describe the workloads, compute platforms, data collection procedures, and collected data.

\subsection{Platforms}
\label{appendix:cluster}

Each platform in our dataset consists of a (device, runtime) tuple. While datasets could conceivably include additional platform dimensions such as the operating system, scheduler, and CPU frequency governor, we chose to study hardware devices and WebAssembly runtimes since these are most relevant to the WebAssembly community.

\paragraph{Devices}

Our cluster (shown in Figure~\ref{fig:cluster}) includes 24 devices from 9 different vendors (Intel, AMD, SiFive, Broadcom, NXP, Amlogic, RockChip, Allwinner, STMicroelectronics) across 14 different microarchitectures (Table~\ref{tab:cluster}). Notable devices include the RISC-V-based Starfive VF2 and the Cortex-M7-based Nucleo-F767ZI.

\paragraph{Runtimes}

For each device, we ran 5 different WebAssembly runtimes with a total of 10 different configurations, including interpreted, ahead-of-time compiled (AOT), and just-in-time compiled (JIT) runtimes (Table~\ref{tab:wasm_runtimes}). Each runtime was run on each device except where not supported: only AOT WAMR runs on the cortex M7, and only WAMR and wasm3 run on the RISC-V device. Ahead-of-time-compiled WAMR was also excluded from Cortex A-72-based platforms due to a code generation bug which can randomly cause illegal instruction errors.

\begin{table}
    \caption{WebAssembly runtimes used. WAMR (the \textit{WebAssembly Micro Runtime}) is also commonly referred to as ``iwasm''.}
    \vspace{0.5em}
    \centering
    {\small
    \begin{tabular}{c p{2.1in}}
        \toprule
        Runtime & Runtime Type \\
        \toprule
        Wasm3 & Interpreter \\
        WAMR & Interpreter, LLVM AOT \\
        WasmEdge & Interpreter \\
        Wasmtime & Cranelift AOT, Cranelift JIT \\
        Wasmer & Singlepass JIT, Cranelift JIT, Cranelift AOT, LLVM AOT \\
        \toprule
    \end{tabular}
    }
\label{tab:wasm_runtimes}\end{table}

\subsection{Side Information}
\label{appendix:side_information}

\begin{figure*}[t]
\centering
\begin{subfigure}[t]{0.49\textwidth}
    \centering
    \includegraphics[width=\columnwidth]{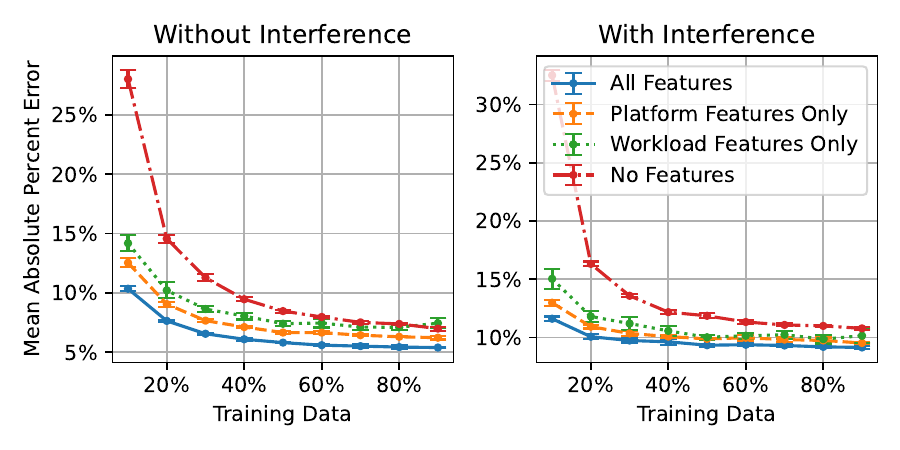}
    \caption{Uncropped version of Figure~\ref{fig:ablation_features}; removing both workload and platform features from Pitot leads to much higher error when only a small amount of data is observed.}
    \label{fig:ablation_features_full}
\end{subfigure}
\hfill
\begin{subfigure}[t]{0.49\textwidth}
    \centering
    \includegraphics[width=\columnwidth]{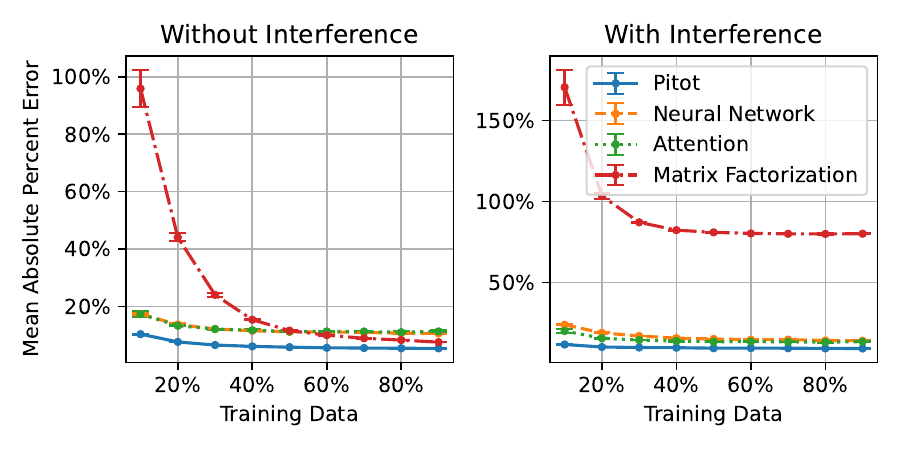}
    \caption{Uncropped version of Figure~\ref{fig:baseline_error}; the Matrix Factorization baseline performs an order of magnitude worse with less training data and predicting interference (since it is not interference-aware).}
    \label{fig:baseline_error_full}
\end{subfigure}
\caption{Uncropped versions of figures where the y-axis was cropped for clarity.}
\label{fig:uncropped_figures}
\end{figure*}

\paragraph{Workload Features}

In order to collect the ``opcode count'' (the number of times each opcode was executed) for each workload, we instrumented the WebAssembly Micro Runtime (WAMR) fast interpreter \citep{wamr-fast} to increment an opcode counter table each time each instruction was executed. Due to several order-of-magnitude differences in opcode counts between short and long benchmarks as well as rare and common instructions, we transform the opcode counts by the log-frequency $f(n) = \log(n + 1)$ (so that $f(0) = 0$). We also exclude opcodes which are not used by any of the workloads from the dataset.

While it is possible to reduce this profiling overhead through an instrumentation-based opcode counting approach, profiling of any kind at this level of detail will be expensive relative to execution without any profiling. However, profiling does not need to be performed on the edge: opcode frequency does not depend on the underlying hardware and only needs to be performed once. As such, profiling can use a fast computer before a workload is to be deployed or is observed for the first time, and does not need to be run during deployment (in the case of edge orchestration) or on a highly-constrained candidate edge device (in the case of system design).

\paragraph{Platform Features}

In addition to a one-hot encoding of the WebAssembly runtime used, we recorded a number of features via linux \texttt{cpuinfo} and \texttt{meminfo}:
\begin{itemize}[topsep=0pt, itemsep=0pt]
    \item CPU microarchitecture (e.g. \texttt{znver3}, \texttt{cortex-a72}, \texttt{tigerlake}), which is one-hot encoded.
    \item Nominal CPU Frequency (i.e. differently clocked CPUs with the same microarchitecture). Note that clock frequency governors (e.g. \texttt{ondemand}) may set the CPU frequency on-the-fly in a highly dynamic manner, which we cannot easily record.
    \item Memory architecture: L1d / L1i cache sizes, L2 size, L2 line size and associativity, L3 size, and main memory size. Cache sizes are passed as a log size, while line size and associativity are provided as one-hot features. Each cache feature is augmented with an indicator feature to account for cases where a given level in the memory hierarchy is not present (e.g. the ARM Cortex-A72 architecture does not have a L3 cache).
\end{itemize}

\subsection{Collected Data}
\label{appendix:data}

\paragraph{Benchmarks in Isolation}

We ran each benchmark on each of our (device, runtime) platforms where supported. In total, we collected 53,637 observations of valid (workload, platform) pairs, and recorded the wall clock execution time for each, averaged over up to 50 repeated executions over a maximum of 30 seconds. While we attempted to execute every possible (workload, platform) pair, some combinations resulted in errors or crashes, which we omit from the dataset. Notable omissions include some WebAssembly runtimes lacking full ARM and RISC-V support at present, interpreted runtimes struggling to complete large benchmarks before being timed out (especially on slower devices), and various implementation bugs on some combinations of runtimes, platforms, and benchmarks.

\paragraph{Interference Dataset}

To evaluate our interference model, we also ran up to 4 benchmarks simultaneously. Each benchmark was run continuously in a loop, resulting in random program alignments. In total, we collected 357,333 usable observations, which includes 98,957 observations with two simultaneously running workloads, 139,208 with three simultaneously running, and 119,168 with three simultaneously running.

During interference data collection, we ran 250 random sets of 2, 3, and 4 workloads on each platform (for a total of 750 sets). Each workload was run repeatedly for 30 seconds. If any of the workloads in a set crashed or otherwise terminated before the end of the 30-second period, that entire set was excluded. Workloads which timed out and failed to complete by the end of the 30-second period but did not crash were also excluded, though other simultaneously running workloads in that set were still included in the dataset since timed-out workloads still cause interference.

\section{Additional Results}

\subsection{Uncropped Figures}
\label{appendix:uncropped}

Figure~\ref{fig:ablation_features} and figure~\ref{fig:baseline_error} were cropped in the y-axis for clarity; we provide uncropped versions of these figures in Figure~\ref{fig:ablation_features_full} and Figure~\ref{fig:baseline_error_full}, respectively.

\subsection{Hyperparameter Ablations}
\label{appendix:hyperparameters}

 \begin{figure*}
    \centering
    \includegraphics[width=1.0\textwidth]{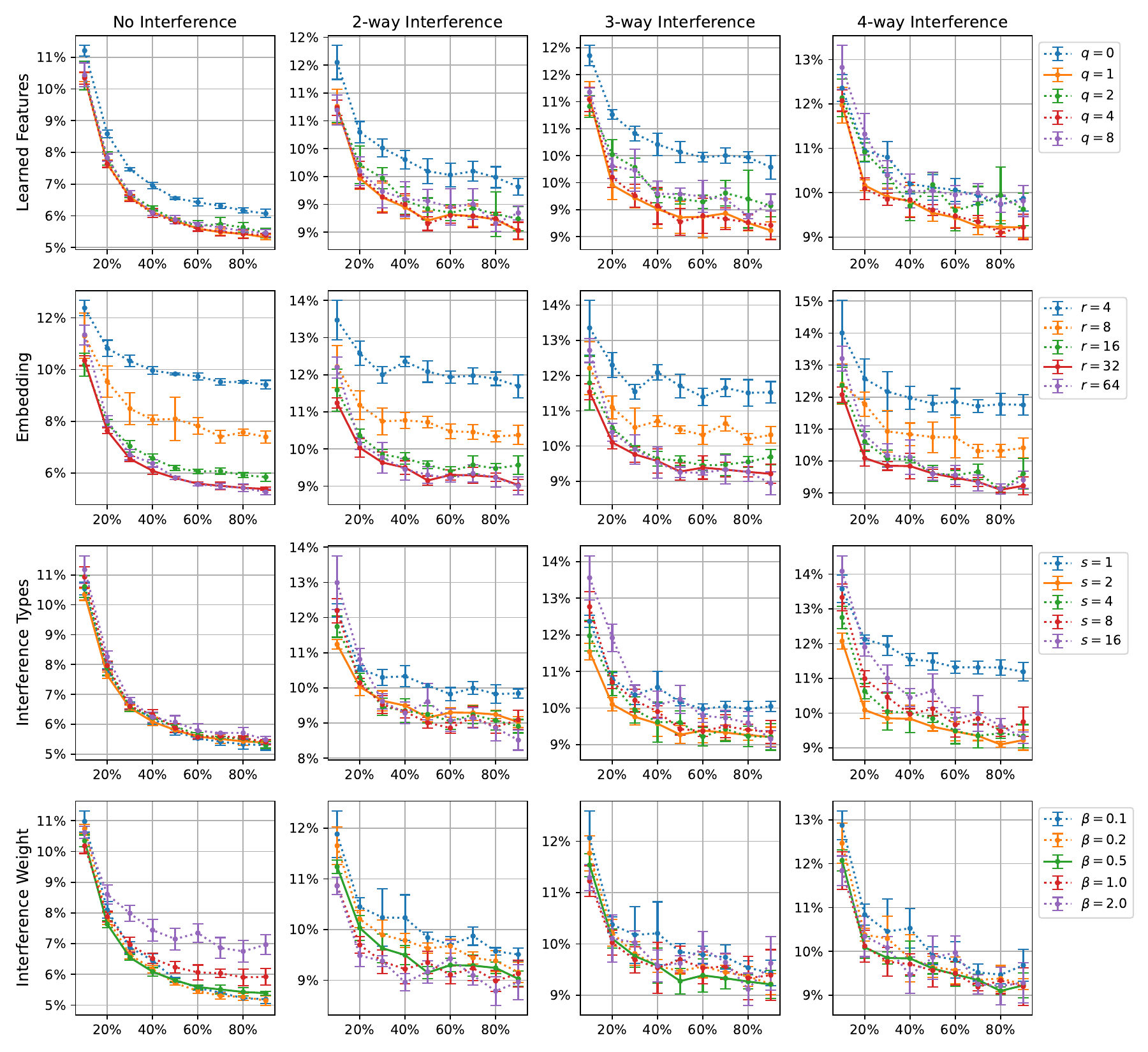}
    \caption{Hyperparameter ablations for the number of learned features, embedding dimension, interference types, and interference objective weight, with mean absolute percent error on the y-axis, and the proportion of observed data on the x-axis. We split our results in each column depending on the number of simultaneously running workloads due to the increased prediction error (and intrinsic problem difficulty) associated with more interfering workloads. In each plot, the solid line indicates the selected hyperparameter value; error bars indicate $\pm 2$ standard errors.}
    \label{fig:hyperparameters}
\end{figure*}

\begin{figure*}[t]
    \centering
    \begin{subfigure}[t]{0.5\textwidth}
        \centering
        \includegraphics[width=\textwidth]{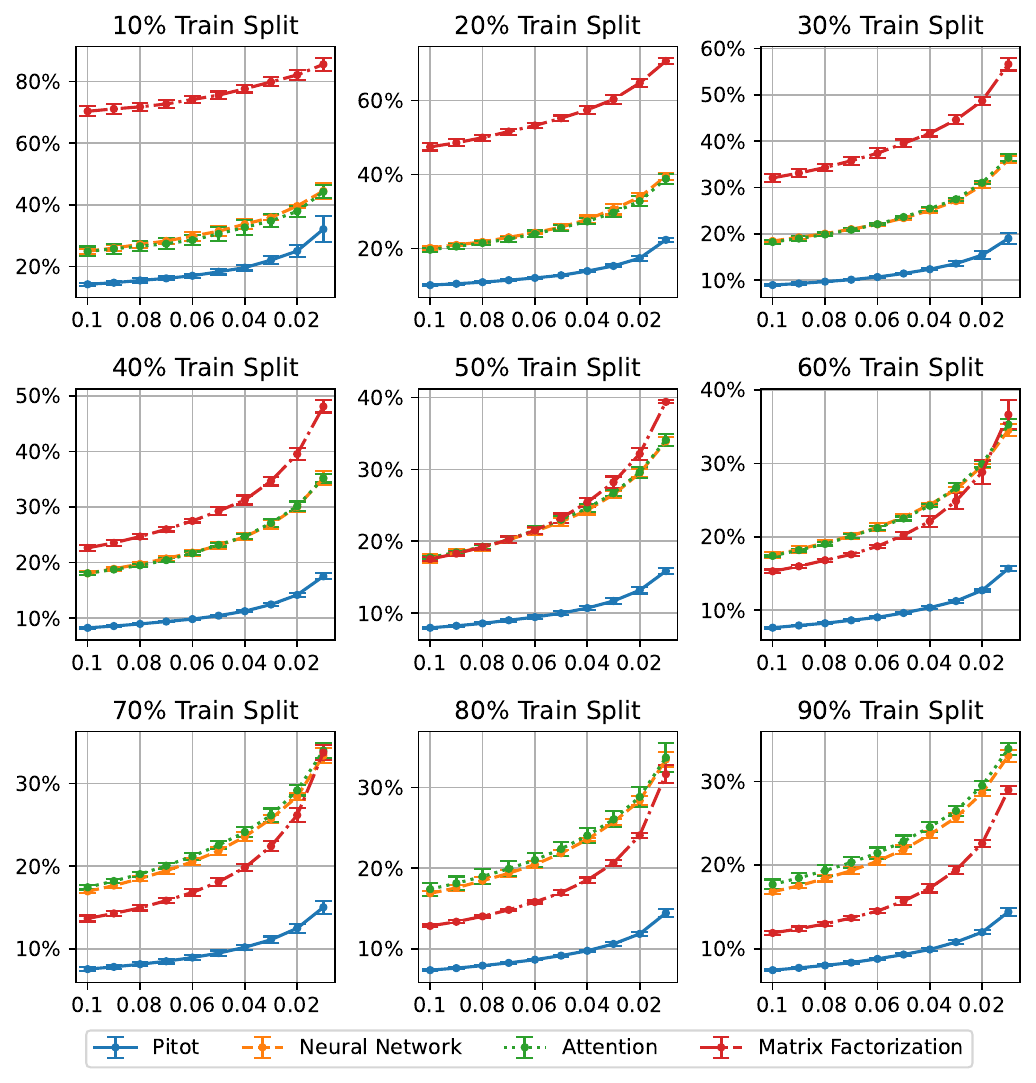}
        \caption{Bounds for prediction without interference}
    \end{subfigure}\begin{subfigure}[t]{0.5\textwidth}
        \centering
        \includegraphics[width=\textwidth]{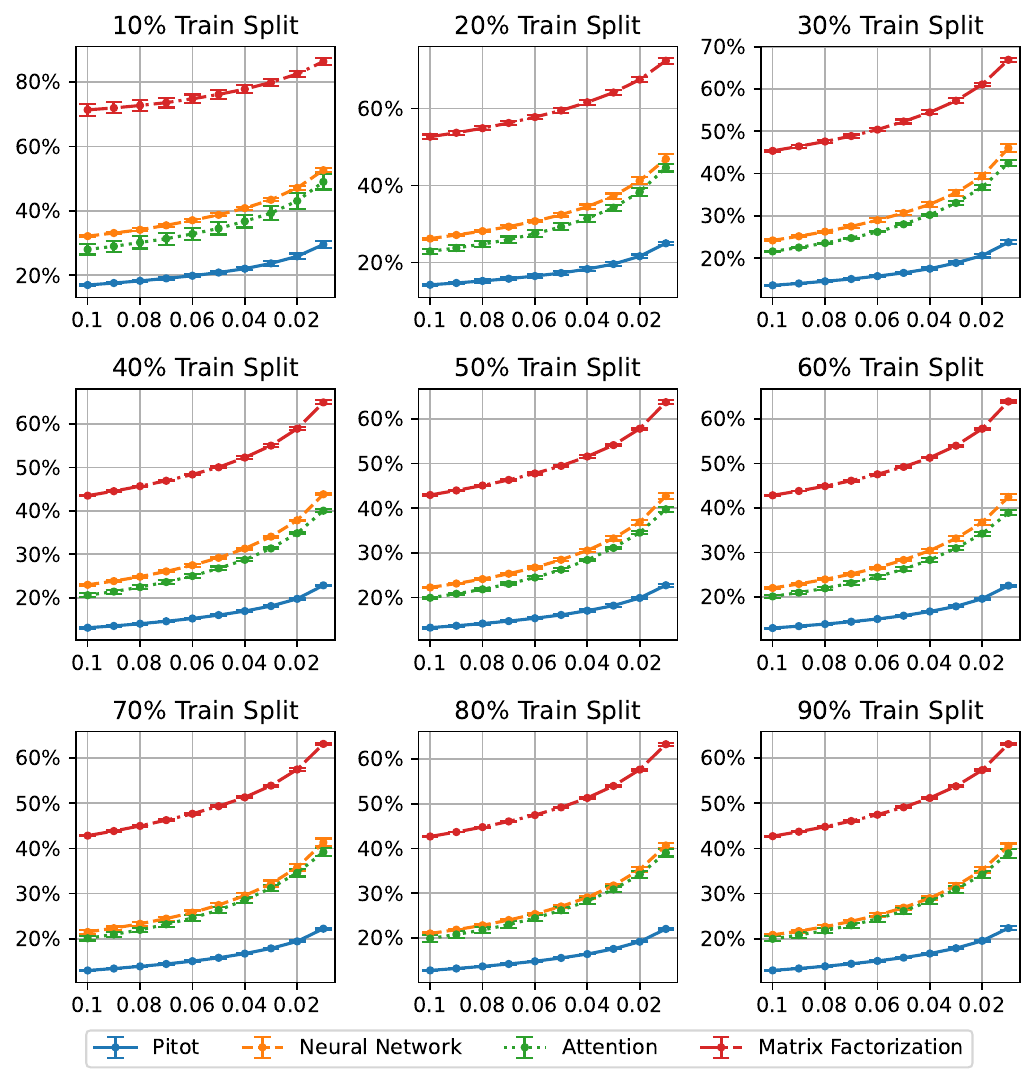}
        \caption{Bounds for interference prediction}
    \end{subfigure}
    \caption{Full bound tightness comparison between Pitot and baselines for the conformal prediction task across varying amounts of training data; each plot shows the bound tightness (with $\pm 2$ standard errors) for a given training split size and varying miscoverage rates.}
    \label{fig:baseline_width_full}
\end{figure*}

We conducted ablation studies on four key hyperparameters for our method, and ran an exponential spread of 5 different values for each (Figure~\ref{fig:hyperparameters}). Our method is not sensitive to the parameters selected, and will perform close to optimally as long as the model embeddings have sufficient dimensionality and thus representational power.

\paragraph{Number of Learned Features $q$}

Learned features in Pitot are feature vectors associated with each platform and workload, which are jointly trained with the embedding network parameters using gradient descent. There is a significant decrease in error in all categories after introducing just one additional learned feature, indicating the necessity of this aspect of Pitot. However, adding additional features does not make a significant impact on model performance. We select $q=1$ for our experiments; in general, any small value of $q$ should be sufficient, though higher $q$ may be beneficial for larger datasets.

\paragraph{Embedding Dimension $r$}

The embedding dimension is the output dimensionality of Pitot's workload and platform embedding networks, and acts as the rank constraint for matrix factorization. In our ablations, we can see a significant improvement in error as the dimensionality increases up to 32 dimensions, after which the error no longer improves. We select $r=32$; in general, $r$ only needs to be sufficiently large, with no significant prediction error downside to an overly large $r$.

\paragraph{Interference Types $s$}

We find that using $s=2$ interference types is sufficient to obtain optimal performance. Our model is slightly sensitive to $s$, with a slight increase in error as $s$ increases for some evaluation settings.

Note that the choice of $s$ does not impact the error of Pitot when predicting the runtime of workloads without any background interference, which is expected, since the interference susceptibility and magnitude embeddings $\bm{v}_s, \bm{v}_g$ are ignored when no interference is present.

\paragraph{Interference Weight $\beta$}

Since Pitot solves a multi-objective optimization problem (even before considering quantile regression), the weight of each objective can impact its error. We assign a constant weight of 1.0 to objectives predicting runtime without interfering workloads, and a weight of $\beta$ to interference prediction, split equally across 2, 3, and 4-way interference.

Increasing the interference objective weight $\beta$ reduces interference prediction error at the cost of increasing error for prediction without interference, with a similar effect in reverse. We choose $\beta=0.5$ as a compromise which does not significantly increase the prediction error for either objective.

\begin{figure*}[t]
    \centering
    \begin{subfigure}[t]{0.5\textwidth}
        \centering
        \includegraphics[width=\textwidth]{images/tsne_workload.pdf}
        \vspace{-2em}
        \caption{Workload features (t-SNE) sorted by benchmark suite}
        \label{fig:tsne_workload}
    \end{subfigure}\begin{subfigure}[t]{0.5\textwidth}
        \centering
        \includegraphics[width=\textwidth]{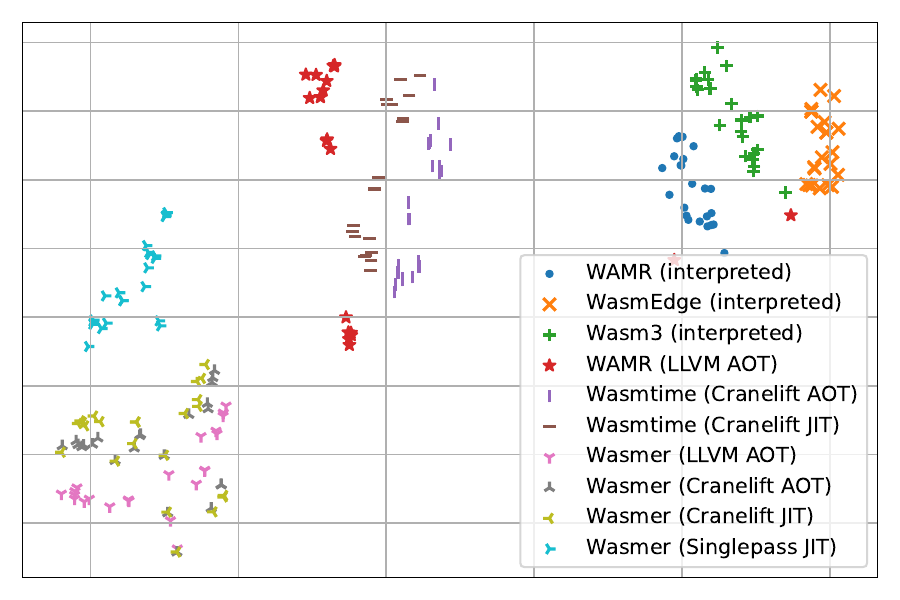}
        \vspace{-2em}
        \caption{Platform features (t-SNE) sorted by WebAssembly runtime}
        \label{fig:tsne_runtime}
    \end{subfigure}

    \begin{subfigure}[t]{0.5\textwidth}
        \centering
        \includegraphics[width=\textwidth]{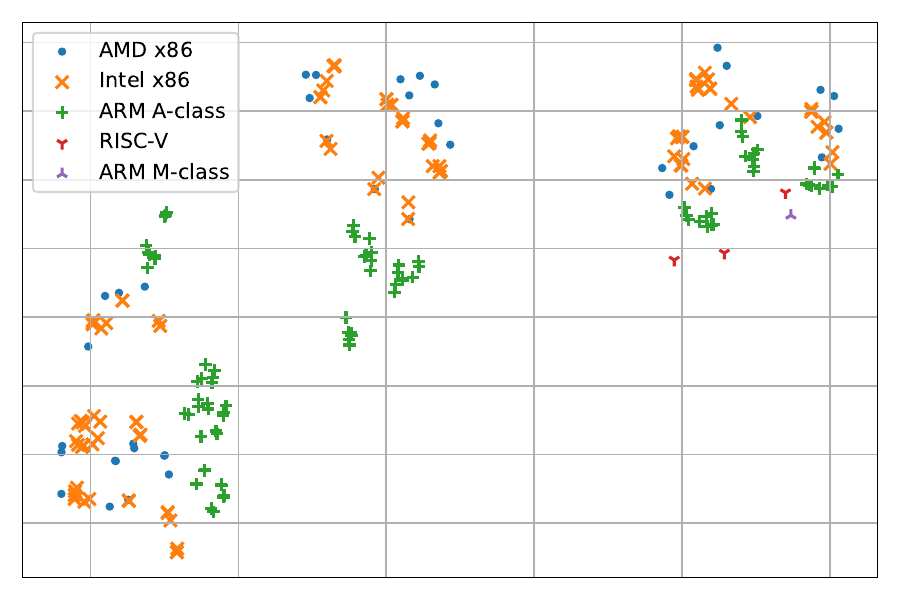}
        \vspace{-2em}
        \caption{Platform features (t-SNE) sorted by CPU microarchitecture}
        \label{fig:tsne_device}
    \end{subfigure}\begin{subfigure}[t]{0.5\textwidth}
        \centering
        \includegraphics[width=\textwidth]{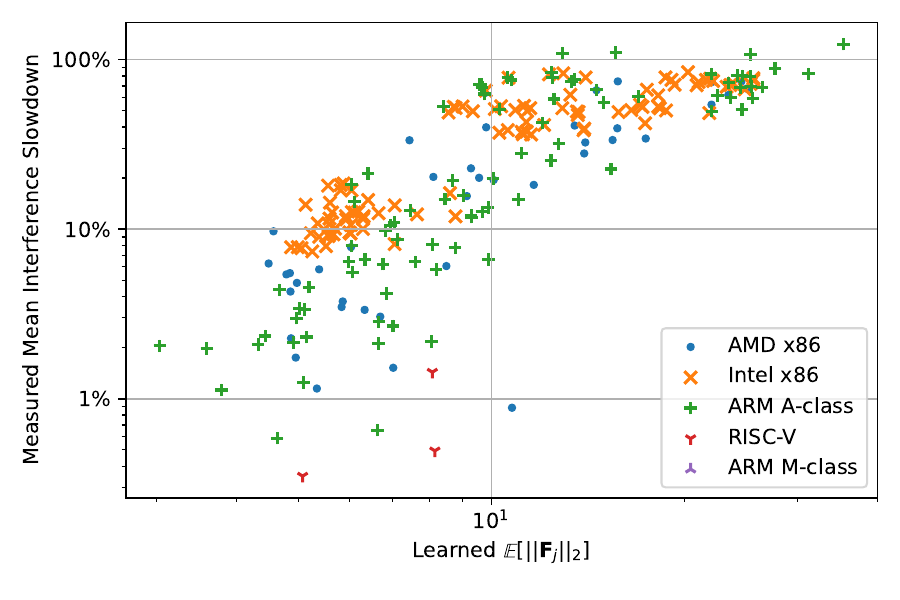}
        \vspace{-2em}
        \caption{Measured and learned interference}
        \label{fig:interference_norm}
    \end{subfigure}
    \caption{Full-size visualizations of Pitot's learned embeddings. Figure~\ref{fig:tsne_workload}-\ref{fig:tsne_device} show t-SNE embeddings of the learned workload and platform features, while Figure~\ref{fig:interference_norm} shows the $l_2$ norm of the learned interference matrix compared with the observed mean interference, sorted by CPU microarchitecture.}
    \label{fig:scatter_full}
\end{figure*}

\subsection{Bound Tightness Comparisons}
\label{appendix:baseline_width_full}

Figure~\ref{fig:baseline_width_full} provides an expanded version of Figure~\ref{fig:baseline_width} showing miscoverage rate-bound tightness curves for each training split size. Pitot performs far better than all of our baselines in each setting, while the attention baseline performs slightly better on predicting interference as the neural network baseline. The matrix factorization baseline performs far worse in most settings, except when predicting runtime without interference when a large proportion of the dataset is observed.

\subsection{Embedding Visualizations}
\label{appendix:visualizations}

Unlike black-box models, Pitot learns embedding vectors for each workload and platform which can be interpreted, and potentially used as inputs for downstream tasks such as anomaly detection. To demonstrate the information value of these learned embeddings, we visualized them by projecting them to two dimensions (Fig.~\ref{fig:tsne_workload}-\ref{fig:tsne_device}). We also analyze our learned interference representation as a sanity check on our interference model (Fig.~\ref{fig:interference_norm}).

\paragraph{Workload Features} To analyze embedding features (Fig.~\ref{fig:tsne_workload}), we project them to 2 dimensions using a t-distributed stochastic neighbor embedding (t-SNE), which maps similar workloads to nearby locations in a 2-dimensional scatter plot, though distances and units do not have any particular meaning. Relatively homogenous benchmark suites such as Polybench, Libsodium, and our Python benchmarks form clear clusters, while more diverse benchmark suites (Mibench, SDVBS, Cortex) are largely mixed.

\paragraph{Platform Features} We also projected platform features using a 2-dimension t-SNE. Sorting platforms by WebAssembly runtime (Fig.~\ref{fig:tsne_runtime}), we see that most runtimes form clear clusters. Notably, the three interpreted runtimes in our dataset (WAMR, WasmEdge, Wasm3) form nearbly clusters, while different configurations of Wasmtime and Wasmer are also respectively clustered together.

Alternatively, organizing platform embeddings by CPU microarchitecture (Fig.~\ref{fig:tsne_device}), we also see clear clusters of each microarchitecture category within the larger clusters associated with each type of runtime.

\paragraph{Interference Matrix} The interference matrix $\bm{F}_j$ allows us to gain insight into the interference characteristics of each platform. Specifically, consider the spectral norm $||\bm{F}_j||_2$,
\begin{align}
    ||\bm{F}_j||_2^2 = \sup_{||w_i||_2 = 1, ||w_k||_2 = 1} \bm{w}_i^T\bm{F}_j \bm{w}_k.
\end{align}
This can be interpreted as the maximum possible interference between two workloads $\bm{w}_i, \bm{w}_k$. Figure~\ref{fig:interference_norm} shows the spectral norm of $F_j$ (trained on the 90\% data split and average over the 5 replicates) plotted against mean interference on each platform. We observe a positive correlation between $||\bm{F}_j||_2$ and measured interference on each device, as we would expect from our interpretation.

\end{document}